\documentclass[11pt]{article}

\usepackage[final]{acl}

\usepackage{times}
\usepackage{latexsym}
\usepackage{amsmath}
\usepackage{amssymb}
\usepackage{tabularx}
\usepackage{makecell}
\usepackage{xspace}

\usepackage[T1]{fontenc}
\usepackage[utf8]{inputenc}

\usepackage{microtype}

\usepackage{inconsolata}

\usepackage{graphicx}
\usepackage{booktabs}

\newcommand{\eg}{e.g.,\xspace}

\newcommand{\ci}[2]{{\scriptsize\,[#1,\,#2]}}

\title{Measuring the Behavioral Fidelity of Long-Horizon Human Activity Simulations}

\author{
    Yi Fei Cheng\textsuperscript{1},
    Fan Yang\textsuperscript{2},
    Iremsu Bas\textsuperscript{1},
    Koichiro Niinuma\textsuperscript{3},
    Narishige Abe\textsuperscript{2}, 
    David Lindlbauer\textsuperscript{1}
\\
\small
\textsuperscript{1}Human-Computer Interaction Institute, School of Computer Science, Carnegie Mellon University, Pittsburgh, PA, USA
\\
\small
\textsuperscript{2}Fujitsu Ltd., Kawasaki, Japan
\\
\small
\textsuperscript{3}Fujitsu Research of America, Pittsburgh, PA, USA
\\
\small
\texttt{\{yifeic2,ibas\}@andrew.cmu.edu} 
\hspace{0.5em}
\texttt{\{fan.yang,kniinuma,abe.narishige\}@fujitsu.com} 
\hspace{0.5em}
\texttt{davidlindlbauer@cmu.edu}
}
\begin{document}
\maketitle

\begin{abstract}
As LLM-based human simulators are increasingly used for policy, evaluation, and training, they must faithfully reproduce real behavioral patterns. While prior work has examined behavioral fidelity in survey responses and dialogue, longer-horizon real-world activity remains largely unexplored. We introduce a framework for evaluating behavioral fidelity in long-horizon activity simulations across temporal granularities and levels of analysis. As a case study, we collect a 43-hour multi-camera dataset of in-the-wild office activity and compare trace-derived conditioning mechanisms: persona descriptors, few-shot exemplars, and statistical transition and time-of-day priors. We find that behavioral fidelity is not uniform across metrics: statistical priors bring activity and sequence distributions closest to real behavior, yet over-fragment routines and suppress within-person variability. These findings motivate a more holistic evaluation that spans multiple metrics, temporal granularities, and levels of analysis.
\end{abstract}

\section{Introduction}
Large language models are increasingly being used to simulate human behavior. 
They have been applied to replicating survey responses~\cite{kolluri2025finetuning,park2026llmagentsgroundedselfreports}, generating human-like dialogues~\cite{suh2026quantifyingusersimulators,gandhi2026humandialogue}, and modeling daily activity sequences~\cite{li2025xsday,park2023generativeagents}. 
These capabilities have enabled a growing range of applications, from hypothesis exploration for policy~\cite{li2026whatif} to interactive evaluation of AI assistants~\cite{yao2024taubench}.

Underlying these applications is the assumption that LLM-generated behavior faithfully reflects real behavioral patterns. 
However, LLMs can vary substantially in their \textbf{behavioral fidelity}. 
In survey questionnaires, LLMs often fail to exhibit human-like response biases~\cite{tjuatja2024llmhuman}. 
In dialogue settings, they can be excessively cooperative~\cite{zhou2026mindsim2realgapuser}.
These limitations undermine the utility of LLM-based simulations in their promised downstream applications.

While a growing body of work investigates the behavioral fidelity of simulated survey responses and dialogue, behavioral fidelity in longer-horizon settings remains largely unexplored~\cite{park2023generativeagents,bougie2025citysim,li2025xsday}.
Recent work has begun to examine the ability of LLMs to simulate long-term digital interaction traces~\cite{chen2026omnibehavior}.
We build on this line of work by evaluating the behavioral fidelity of LLMs in simulating \textbf{long-horizon real-world human activities}.
We refer to long-horizon activities as those that span multiple hours or days (\eg~a day in the life of a knowledge worker, including arriving at the office, working on a computer, getting coffee, and engaging in conversations).

In this work, we ask: \textbf{do simulated agents reproduce the everyday activity patterns of real people over long time horizons?}
To address this question, we introduce a framework for systematically evaluating the behavioral fidelity of long-horizon human activity simulations.
We decompose behavioral fidelity along two dimensions: \textbf{temporal granularity} (activity-level, time-of-day, and day-level fidelity) and \textbf{level of analysis} (individual- and population-level fidelity).

To demonstrate the applicability of our framework, we collected a 43-hour multi-camera office dataset capturing the in-the-wild activities of 55 people, five of whom were each observed for $4\pm1$ days, for $7\pm1$ hours per day.
Using this dataset, we evaluate the behavioral fidelity of six simulation approaches, varying in how they incorporated real-world behavioral traces as priors, including a no-persona control, authored and trace-inferred persona descriptors, few-shot full-day behavioral exemplars, and statistical priors over activity transitions and time-of-day prevalence.

Our results indicate that, in our experimental setting, statistical priors improve alignment
with real activity and sequence distributions, but at the cost of over-segmented
days and reduced within-person variability. 
We also find that differences among persona-based conditioning approaches are small by comparison, and depend on which metric is used to assess them.
Finally, we observe that population-level agreement can mask errors at the
individual level. 
Evaluating simulators across multiple temporal granularities and levels of analysis is therefore essential for characterizing their strengths and limitations, an evaluation that our framework is designed to support.

\paragraph{Contributions.} Our contributions are threefold:
\vspace{-.5em}
\begin{enumerate}
    \setlength{\itemsep}{-.5em}
    \item A framework for measuring the behavioral fidelity of long-horizon human activity simulators across multiple temporal granularities and levels of analysis.
    \item A 43-hour multi-camera dataset of in-the-wild office activities, enabling evaluation of simulated long-horizon activities against real-world activity traces.\footnote{Our dataset and supplementary materials are available at \url{https://augmented-perception.org/publications/2026-long_horizon_agents.html}}
    \item A case study applying the framework to six conditioning approaches, showing that behavioral fidelity is non-uniform across metrics, temporal granularities, and levels of analysis.
\end{enumerate}

\section{Related Work}
\label{sec:related}

\subsection{LLM-Based Agents}

In recent years, substantial research has focused on extending LLMs into LLM-based agents~\cite{wang2024survey}. 
By scaffolding LLM generation with components such as memory structures and planning modules, these systems have demonstrated the ability to perform complex multi-step tasks across diverse domains without task-specific fine-tuning~\cite{yao2023react,shinn2023reflexion}.
Most relevant to our work are recent efforts using LLM-based agents to simulate human behavior.

\citet{park2023generativeagents} first introduced \emph{generative agents} that leveraged LLMs in a perceive--reflect--plan--act loop to simulate believable everyday activities and social interactions in a simulated environment.
Subsequent work has extended this paradigm in several dimensions. 
One line of research focused on modeling richer internal human characteristics, such as needs, emotion, and personality traits~\cite{wang2025d2a,wang2023humanoid,liang2025actor,li2025xsday}.
A complementary line of work emphasized physical grounding and embodiment, introducing mechanisms to condition and constrain generated behaviors based on environmental context~\cite{liang2025actor,wu2025indoorworld,li2025xsday}. 
Finally, \citet{bougie2025citysim} and \citet{piao2026agentsociety} explored scaling these simulations to large multi-agent societies.

Existing approaches have shown promise in generating ``believable'' human behavior~\cite{park2023generativeagents}. 
However, for downstream applications such as policy evaluation~\cite{li2026whatif}, 
simulations must not only appear plausible,
but also faithfully reproduce real human behavioral patterns.
Our research builds on this literature by introducing a framework to systematically evaluate the fidelity of these behaviors.

\subsection{Measuring the Behavioral Fidelity of Human Simulations}

As LLMs are increasingly used to simulate humans, a natural question is how
faithfully they do so.
Prior work has shown that LLMs often fail to reproduce human behavioral patterns
in surveys and opinion
studies~\cite{tjuatja2024llmhuman,bisbee2024syntheticreplacement}.
Studies of dialogue simulation have identified discrepancies in conversational
style, engagement patterns, problem-solving strategies, emotional expression,
and behavioral diversity between LLM-simulated and real human
interactions~\cite{ivey2024realroboticassessingllms,wang2025humanvsagent,zhou2026mindsim2realgapuser}.
SimBench~\cite{hu2026simbench} recently aggregated tasks ranging from moral
decision-making to economic choice, and found the same pattern.
Collectively, this body of work shares the objective of assessing simulation
\emph{fidelity}: the extent to which a model reproduces the response
distribution of the population it is conditioned on~\cite{argyle2023outofone}.

While our work shares this objective, we focus on open-ended, long-horizon
behavioral patterns rather than single-turn, self-contained
questions~\cite{hu2026simbench} or dialogue~\cite{mehri2026distributionalgap}.
In this setting, \citet{li2025digitaltwins} demonstrated that accurately
reproducing continuous behavioral trajectories remains a significant challenge
for LLMs, and \citet{chen2026omnibehavior} found that simulations of long-term
digital interaction traces tend to converge toward a positively skewed
``average person'' representation.
In contrast to personas and user simulators derived from digital interaction
logs~\cite{chen2026omnibehavior} or fictional and biographical
sources~\cite{li2025digitaltwins}, our work grounds simulation in traces of
real-world physical activity.

\section{Long-Horizon Activity Simulation}
\label{sec:problem}

Our work focuses on evaluating the behavioral fidelity of LLM-based simulators of long-horizon human activity. 
We build on the generation problem introduced by \citet{wang2025d2a}.

An LLM-based long-horizon human activity simulator is an agent that
autoregressively generates sequences of activities within an interactive
environment over the course of a day.

Each activity is realized as an \emph{activity episode}
$\sigma = (a, t_{\text{start}}, t_{\text{end}})$: one uninterrupted stretch of a
single activity, labeled $a$ from a closed activity vocabulary $\mathcal{A}$,
running from time $t_{\text{start}}$ to time $t_{\text{end}}$.
Formally, at each step $m$, the agent samples an activity,
\begin{equation}
\label{eq:policy}
    a_m \sim \pi \!\left( \cdot \mid \sigma_{1:m-1},\, o_{1:m-1},\, e,\, p \right),
\end{equation}
conditioned on the episode history $\sigma_{1:m-1}$, prior observations
$o_{1:m-1}$ of environment $e$, and a persona $p$ describing attributes such as
identity, role, routines, and behavioral tendencies.
The sampled activity is executed until it concludes, yielding the episode
$\sigma_m$, which begins where $\sigma_{m-1}$ ended.
Rolling out the policy for $M$ steps yields an \emph{activity trajectory}
$\tau = (\sigma_1, \ldots, \sigma_M)$, a contiguous sequence of episodes
spanning the simulated day.
\begin{table*}[t]
\centering
\footnotesize
\setlength{\tabcolsep}{5pt}
\begin{tabular}{@{}p{0.18\linewidth}p{0.60\linewidth}@{}}
\toprule
\textbf{Temporal granularity} & \textbf{Behavioral metrics} \\
\midrule

\textbf{Local} &
Episode durations, transitions, behavioral motifs (\eg~3-gram, 4-gram) \\[4pt]

\textbf{Time-of-day} &
Time allocation, frequency, switches \\[4pt]

\textbf{Day} &
Time allocation, frequency, switches, \mbox{within-person variability (individual)}, \mbox{between-person variability (population)} \\

\bottomrule
\end{tabular}
\caption{
Behavioral fidelity metrics organized by temporal granularity.
Each metric is evaluated at both the individual and population levels of analysis unless otherwise specified.
}
\label{tab:taxonomy}
\end{table*}

\section{Evaluating Behavioral Fidelity}
\label{sec:framework}
We define \textbf{behavioral fidelity} as the extent to which simulated activity trajectories reproduce patterns observed in real human activity traces.

Human behavior exhibits structure across multiple temporal and population scales.
Some regularities emerge only at coarse temporal resolutions, such as overall activity time allocation throughout the day, while others depend on finer-grained sequential dynamics, such as activity transitions and recurring routines.
Likewise, fidelity can differ across levels of analysis: a simulator may reproduce aggregate population statistics while failing to preserve individual-specific tendencies, or generate plausible individual trajectories that collectively deviate from real population behavior.
Therefore, our proposed framework evaluates the behavioral fidelity of long-horizon human activity simulators along two dimensions: \textbf{temporal granularity} and \textbf{level of analysis}.

\paragraph{Temporal granularity.}
Human activities exhibit structure across multiple temporal granularities.

At the \textbf{local} level, behavior can be characterized by structure within short spans of adjacent activities, including activity \emph{episode durations}, \emph{transitions}, and short-range \emph{behavioral motifs}.
For example, information workers frequently transition between focused work, communication, and other tasks throughout the day~\cite{dabbish2011interrupt}.

At the \textbf{time-of-day} granularity, we compute descriptors within fixed time windows (e.g., 09:00--11:00, 11:00--13:00, \ldots), capturing higher-level behavioral patterns across different times of day.
We characterize this using activity \emph{time allocation} and \emph{frequency}.
Individuals may allocate different portions of the day to different activities, such as computer work in the morning and meetings later in the day, producing recurring daily work rhythms~\cite{hernandez2025triplepeakday}.

Finally, at the \textbf{day} level, behavior can similarly be characterized in terms of overall activity \emph{time allocation} and \emph{frequency}.
For instance, individuals may allocate more time to socializing than others.
Behavior can also be characterized with respect to the number of activity \emph{switches} that occur within a day.
Some individuals may have a tendency to engage in prolonged tasks, while others are easily interrupted.
Across days, individuals may differ in the \emph{variability} of their behaviors, ranging from highly consistent daily routines to substantial day-to-day variation in behavior.

\paragraph{Level of analysis.}
Behavioral patterns can be characterized at the \textbf{individual} and \textbf{population} levels.
At the \textbf{individual} level, fidelity concerns whether a simulator preserves the characteristic tendencies and behavioral styles of specific individuals, such as preferences for particular activities, transition habits, or degrees of routine consistency.
At the \textbf{population} level, fidelity concerns whether simulated behaviors collectively reproduce aggregate statistics and behavioral diversity observed across a population, including overall activity frequencies, temporal allocation patterns, and between-person variability.
A simulator may therefore achieve high fidelity at one level while failing at the other, matching population-level statistics while collapsing individual differences, or producing plausible individual routines that collectively diverge from real-world population behavior.

\subsection{Statistical Framework}
\label{sec:metrics}
Let $\mathcal{P}$ denote the distribution of real behaviors and $\mathcal{Q}$
denote the distribution of simulated behaviors.
The behavioral fidelity of a simulator can thus be defined as the extent to
which $\mathcal{Q}$ approximates $\mathcal{P}$.

We estimate $\mathcal{P}$ empirically using a reference dataset of human
activity trajectories, denoted $\mathcal{D}_{\text{real}}$.
Each trajectory in this dataset is enacted by an individual $i \in \mathcal{I}$
during a session $s \in \mathcal{S}_i$, so that
$\mathcal{D}_{\text{real}} = \{\tau_i^{(s)} : i \in \mathcal{I},\, s \in
\mathcal{S}_i\}$.
We estimate $\mathcal{Q}$ using a set of simulated trajectories
$\mathcal{D}_{\text{sim}} = \{\tilde{\tau}_i^{(s)} : i \in \mathcal{I},\, s \in
\tilde{\mathcal{S}}_i\}$, each obtained by rolling out the simulation policy of
\S\ref{sec:problem} for an agent conditioned on a persona representing
individual $i$.
Hereafter, we use a tilde to denote simulated quantities.

This enables us to quantitatively measure behavioral fidelity by characterizing
$\mathcal{D}_{\text{real}}$ and $\mathcal{D}_{\text{sim}}$ using metrics derived
from our framework and quantifying the differences between them.
Table~\ref{tab:taxonomy} summarizes the metrics instantiated from our framework.
For distributional behavioral representations (\eg~activity time allocation
distributions), we quantify divergence between real and simulated behaviors
using Total Variation (TV), Jensen--Shannon (JS) divergence, and forward and
backward Kullback--Leibler (KL) divergence.
For count- and duration-valued statistics, we use $L_{1}$ distance and signed
bias metrics to quantify the magnitude and direction of simulation errors.
Operational definitions of the behavioral metrics are reported in~\autoref{app:metrics}.
\section{Data Collection}
\label{sec:data}

As a case study for our evaluation framework (\S\ref{sec:framework}), we collected a reference dataset $\mathcal{D}_{\text{real}}$ comprising 43 hours of in-the-wild activity trajectories from occupants of a shared office. 
We use this dataset in our experiments (\S\ref{sec:experiments}) to examine how conditioning approaches affect behavioral fidelity in this setting.
We report the contents and construction procedure of our dataset in Appendices~\ref{app:dataset} and \ref{app:construction}, respectively.

\begin{figure}[b]
    \centering
    \includegraphics[width=1\linewidth]{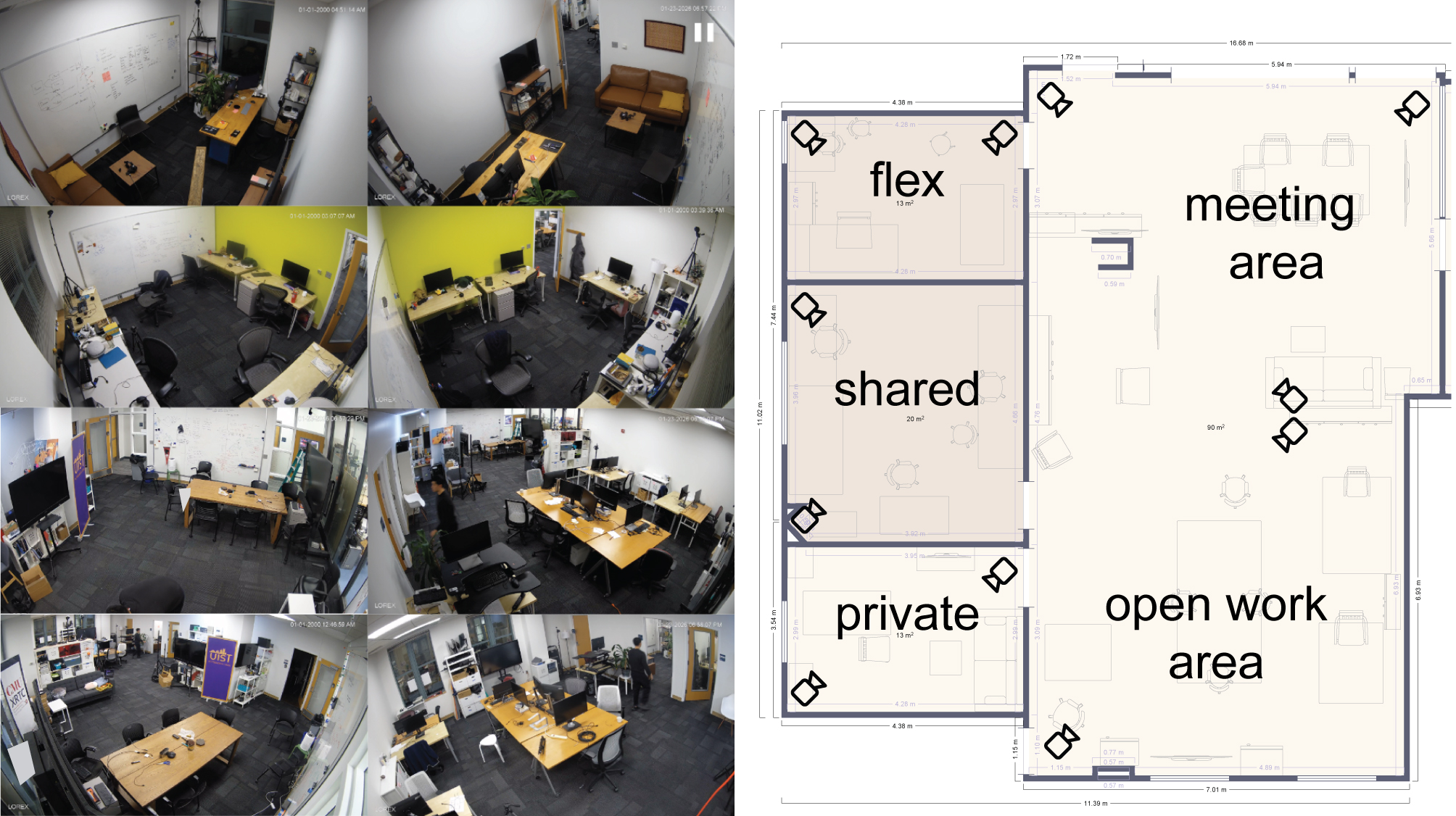}
    \caption{Camera positions used during data collection.}
    \label{fig:apparatus}
\end{figure}

\paragraph{Recording.}
We instrumented a four-room shared office comprising an open work area, shared office, flexible meeting room, and private office with $11$ synchronized cameras that recorded continuously over one work week ($43$ hours of video; shown in \autoref{fig:apparatus}). 
The study was approved by an IRB (STUDY2025\_00000414). 
% occupants were informed and could request removal (none did). We plan to release only derived activity sequences, not raw video or identity-linkable signals.

\paragraph{Activity vocabulary.}
\label{sec:activity-vocabulary}
We instantiate the closed activity vocabulary $\mathcal{A}$ (\S\ref{sec:problem})
with labels covering common office behaviors.
Following \citet{patidar2025organichar}, we derive the label set by
clustering sampled segments and merging overlapping candidates into stable
categories.
This yields six labels: \texttt{use computer},
\texttt{engage in conversation}, \texttt{have meal}, \texttt{take a break},
\texttt{use phone}, and \texttt{out of room}.
This label space is used consistently for annotation and simulation.

\paragraph{Annotation.}
All labels in $\mathcal{D}_{\text{real}}$ are produced with manual annotation
(\autoref{app:construction}).
We converted the recordings into activity trajectories $\tau_i^{(s)}$.
$\mathcal{D}_{\text{real}}$ collects them across all individuals and sessions.

\paragraph{Dataset statistics.}
Our final $\mathcal{D}_{\text{real}}$ contains trajectories from $55$ unique
individuals across $43$ recorded hours. 
Among the captured individuals, five had substantially more representation, averaging $29 \pm 4$~hours across $4 \pm 1$~days.
On days they were present, they were in the office for $7 \pm 1$~hours.
This is in contrast to the remaining $50$ captured individuals (\eg~visitors), who were typically only observed for $1 \pm 1$~day for $1 \pm 1$~hour per day present.
% We centered our analysis on the $N=5$ individuals with significantly more hours, which was a pre-requesite as a long-horizon persona representation.
We focused our analysis on the five individuals with sufficient longitudinal coverage ($4 \pm 1$~days, $7 \pm 1$~hours per day when present) to support long-horizon persona construction. 

\section{Experiments}
\label{sec:experiments}
To demonstrate the framework (\S\ref{sec:framework}) and dataset (\S\ref{sec:data}) in use, we analyze six LLM-based simulation approaches spanning different conditioning mechanisms.
For each method, we generated $\mathcal{D}_{\text{sim}}$ by instantiating a five-agent, eight-hour simulation within an environment designed to mirror the recorded office setting.
Each agent was modeled after one of the five core individuals observed across multiple real-world days in our dataset.
We then quantify the differences between simulated data $\mathcal{D}_{\text{sim}}$ and our real-world reference corpus $\mathcal{D}_{\text{real}}$, and draw conclusions about the efficacy of the
approaches.

\subsection{Environment}
\begin{figure}[t]
    \centering
    \includegraphics[width=1\linewidth]{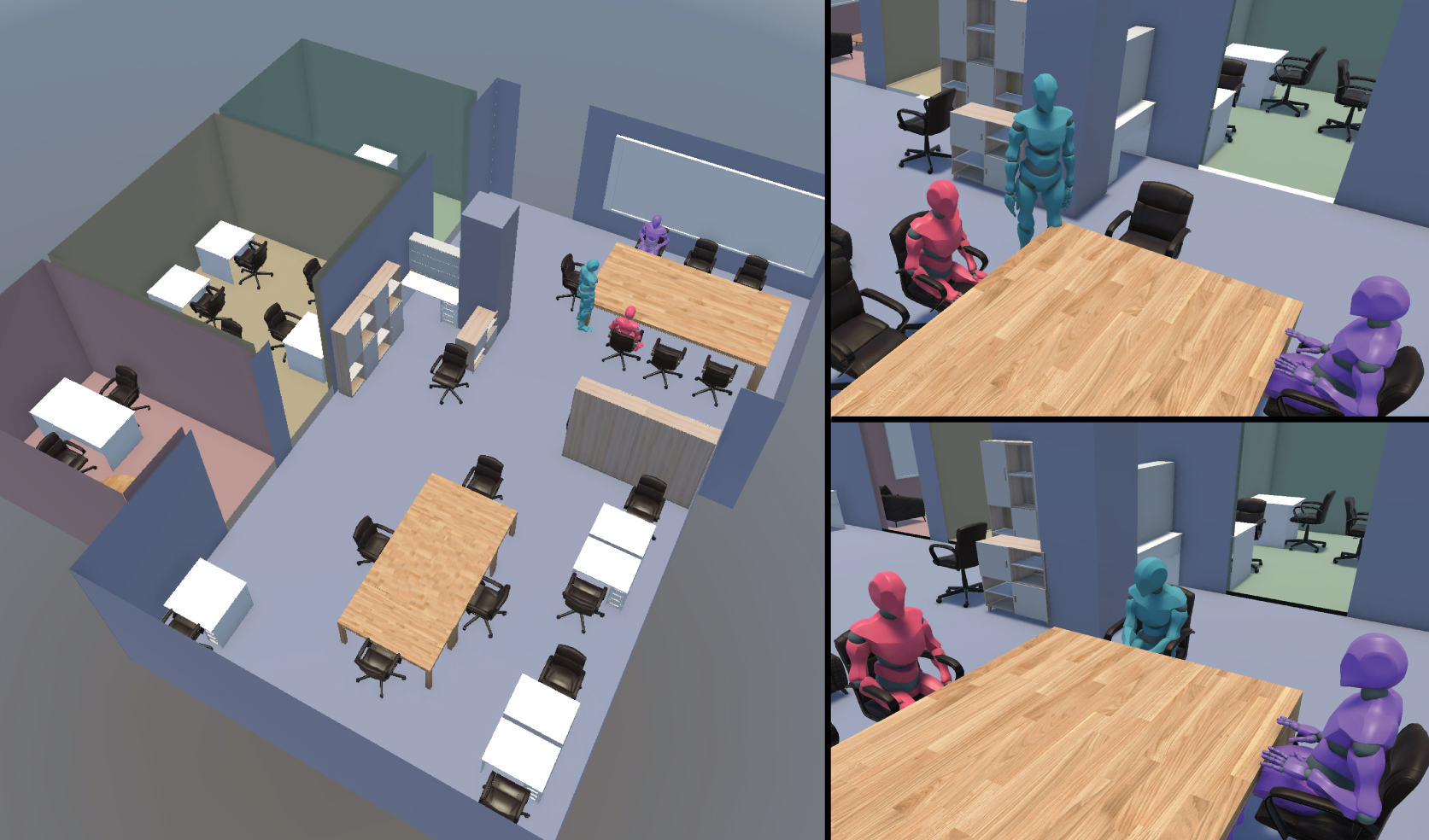}
    \caption{An illustration of the simulation environment used in our experiments. The environment is represented to agents as a scene graph; the 3D rendering is shown for visualization purposes only.}
    \label{fig:environment}
\end{figure}

The simulation environment was designed to mirror the recorded office environment represented in $\mathcal{D}_{\text{real}}$.
Similar to prior work on agent-based simulation~\cite{park2023generativeagents,li2025xsday}, we represent the environment as a hierarchical scene graph capturing rooms, areas, objects, and their spatial relationships.
At each simulation step, the relevant portion of the graph is provided as context to ground agent activity generation.
Figure~\ref{fig:environment} illustrates the environment used in our experiments.

The environment additionally supports multi-agent conversations following \citet{wu2025indoorworld}.
Agents may initiate, join, remain in, or leave conversations, participating in at most one conversation at a time.
Conversation sessions are managed globally within the environment.

\subsection{Agent Architecture}
\label{sec:architecture}

The evaluated simulation approaches all build on a shared agent architecture adapted from recent LLM-based agents~\cite{wu2025indoorworld,park2023generativeagents}.
Our approach integrates multiple cognitive modules within a perceive--plan--act loop: perception, memory, planning, and action.

\paragraph{Initialization.}
Agents are initialized with a structured persona description (\S\ref{sec:conditioning}).
Based on this description and the environment context, the agent generates a \emph{day plan} specifying coarse higher-level intentions over the simulation horizon.
The agent then selects an \emph{activity} and instantiates it as a grounded \emph{action} executed at the current timestep, repeating this at each subsequent timestep until the activity concludes and is realized as an episode spanning one to tens of minutes (\S\ref{sec:problem}).

\paragraph{Perception.}
At each timestep, the perception module gathers observations of the activities of co-located agents and updates the agent's memory accordingly. 
In addition, it identifies interruptions to the agent's current activity based on recent contextual observations.

\paragraph{Memory.}
This module maintains a \emph{working memory}~\cite{baddeley2020working} of the agent's current day plan and activity, and an \emph{episodic memory}~\cite{tulving2002episodic} organized as a two-stage hierarchy~\cite{atkinson1968human}. 
New entries from perception and self-action are first integrated into a \emph{short-term memory} storing ongoing and recently concluded episodes, and are periodically consolidated into a \emph{long-term memory} consisting of condensed summaries spanning fixed temporal windows.

\paragraph{Planning.}
\label{sec:planning}
At episode boundaries or upon interruptions, the planning module uses the agent's observations and memories to update the day plan and determine the next activity.
The day plan is trimmed to the remaining horizon and, in the case of interruptions, revised using the agent's current memories. 
The next activity is then selected following \citet{wang2025d2a}, adapting their activity proposal and evaluation paradigm into a multi-objective candidate sampling and scoring procedure.
This multi-objective formulation allows activity selection to explicitly account for and balance competing behavioral constraints.
As a default configuration, we introduce two equally weighted objective streams. 
One stream samples and evaluates candidate activities based on alignment with the current day plan, while the other samples and evaluates candidate activities based on the agent's immediate context.

\paragraph{Action.}
Finally, the action module integrates information from perception, memory, and planning to select and execute the next action. 
Actions are observable one-timestep behaviors grounded within the environment.

\subsection{Conditioning Approaches}
\label{sec:conditioning}

% ------------------------------------------------------------------
\begin{table*}[t]
\centering
\small
\setlength{\tabcolsep}{4pt}
\begin{tabularx}{\textwidth}{@{}lXccc@{}}
\toprule
\textbf{Variant} &
\textbf{Conditioning setup} &
\makecell{\textbf{LLM}\\\textbf{streams}} &
\makecell{\textbf{Prior}\\\textbf{streams}} &
\makecell{\textbf{Duration}\\\textbf{prior}} \\
\midrule

(C1) \textsc{Authored Persona}
& Persona + Authored Behavior
& \checkmark &  &  \\

(C2) \textsc{Inferred Persona}
& Persona + Trace-inferred Behavior
& \checkmark &  &  \\

(C3) \textsc{Persona + Few-Shot}
& Persona + Example Sequences
& \checkmark &  &  \\

(C4) \textsc{Statistical Prior}
& Persona + Statistical Priors
&  & \checkmark & \checkmark \\

(C5) \textsc{Hybrid Prior}
& Persona + Trace-inferred Behavior + Statistical Priors
& \checkmark & \checkmark & \checkmark \\

(B0) \textsc{None}
& Name + Pronouns
& \checkmark &  &  \\

\bottomrule
\end{tabularx}

\caption{
Conditioning approaches evaluated in our experiments.
LLM streams and prior streams indicate which sources contribute candidate activities and scores during activity selection.
All variants share the same agent architecture, environment, and LLM backbones.
}
\label{tab:variants}
\end{table*}
% ------------------------------------------------------------------

Prior work has explored a range of conditioning approaches for steering the generation of human activity sequences. 
In LLM-based agents, the dominant approach is to prompt a model to role-play as a user profile~\cite{park2023generativeagents,suh2026quantifyingusersimulators}.
In our experiments, we evaluate the behavioral fidelity of six conditioning approaches (Table~\ref{tab:variants}).

\paragraph{(C1) \textsc{Authored Persona}.}
In this approach, simulated agents are conditioned using a structured persona descriptor~\cite{park2023generativeagents} containing their name, pronouns, innate and learned identity, current focus, social relationships, environment-specific context, and behavioral tendencies. 
Agents use the default architecture described in \S\ref{sec:architecture}. 
The persona descriptors were authored by the simulated individuals themselves.

\paragraph{(C2) \textsc{Inferred Persona}.}
This approach is identical to (C1), except that the behavioral tendencies field of the persona descriptor is inferred from the simulated individual's activity traces collected in \S\ref{sec:data}. 
Specifically, we generate natural-language descriptions summarizing behavioral tendencies from sequences of observed activities.

\paragraph{(C3) \textsc{Persona + Few-Shot}.}
This approach is identical to (C1), except that the behavioral tendencies field is excluded from the structured persona descriptor. 
Instead, behavioral grounding is provided through few-shot exemplars consisting of two real full-day activity sequences from our dataset corresponding to the simulated individual.

\paragraph{(C4) \textsc{Statistical Prior}.}
In this approach, agents are conditioned using the structured persona descriptor in (C2--C3), excluding the behavioral tendencies field. 
Within the planning module, the LLM-based candidate sampling and scoring streams (\S\ref{sec:architecture}) are disabled and replaced with trace-derived priors estimated from the observed activity sequences of the corresponding individual. 

Specifically, activity candidates are generated and scored using two prior-based streams. 
A \emph{transition stream} samples candidates from the empirical first-order transition distribution $P(a_{t+1}\mid a_t)$ and scores them using the corresponding transition log-probability conditioned on the current activity. 
A \emph{time-of-day stream} samples candidates from the empirical activity distribution conditioned on the current two-hour workday bin and scores them using the corresponding conditional log-probability. 

Activity labels proposed by either stream are subsequently rendered into natural-language activity descriptions by an LLM, and the resulting prior scores are aggregated within the multi-objective selector. 
Episode durations are sampled from empirical per-activity duration distributions.
The LLM still initializes day plans and executes grounded actions; only episodic activity \emph{selection} is driven by empirical priors.

\paragraph{(C5) \textsc{Hybrid Prior}.}
This approach combines the inferred persona of \textsc{(C2)} with the full four-stream activity selector: LLM plan- and context-based sampling and scoring operate alongside the transition- and time-of-day-prior streams of \textsc{(C4)}. 
% Prior-derived labels are instantiated by the LLM. Durations are drawn from empirical per-activity distributions.

\paragraph{(B0) \textsc{None}.}
As a no-persona control, agents are conditioned only on a name and pronouns. 
\section{Results and Analysis}
\label{sec:results}

We conducted experiments using three LLMs: the open-source Llama 3.3 70B Instruct model, and the proprietary GPT-5.4-mini-2026-03-17 and Gemini 2.5 Flash models.
For each combination of simulation approach and backbone model, we conducted $K=3$ independent simulation runs.

In our work, the objective was to evaluate the behavioral fidelity of long-horizon activity simulations rather than predictive performance on held-out days.
Accordingly, for C2--C5, we derived empirical priors from the full real activity corpus
$\mathcal{D}_{\text{real}}$, including behavioral tendency descriptors (C2), few-shot example sequences (C3), and statistical priors (C4--C5).
We then quantified differences between $\mathcal{D}_{\text{real}}$ and the simulated corpus $\mathcal{D}_{\text{sim}}$ using the metrics described in \S\ref{sec:metrics}, evaluating how different conditioning mechanisms reproduce behavioral patterns when grounded in the same underlying behavioral evidence.

\paragraph{Metrics.}
At the day level, we
report the divergence between the simulated and real sequences with respect to activity \emph{time allocation}, \emph{frequency} distributions, \emph{switches}, and \emph{within-person variability}. 
We report divergences in the activity \emph{time allocation} and \emph{frequency}
distributions within four two-hour windows (09:00--11:00, 11:00--13:00,
13:00--15:00, and 15:00--17:00).
Finally, at a local level, we report divergences in the next-activity \emph{transition}, \emph{3-gram} and
\emph{4-gram} distributions. 
Throughout, unless otherwise specified, divergences are computed at the
individual level (\autoref{app:metrics}).
We report per-approach metrics by averaging across backbones.
We compute 95\% BCa bootstrap confidence intervals ($B = 10{,}000$) for per-approach metrics and paired contrasts by resampling over individuals with replacement; contrasts are Bonferroni-corrected across the $\binom{6}{2} = 15$ approach pairs.
Due to the small number of individuals ($n=5$), these intervals characterize how much each estimate depends on which individuals are included, rather than supporting generalization beyond this cohort.

\paragraph{Summary of Results.}

In this setting, conditioning on statistical priors (C4--C5) reduced
divergence from real behavior on the activity and sequence distributions we
measured (\S\ref{result:stat-activity-sequence}), but the same conditioning generated days with more
activity switches than real ones, and less day-to-day variation within an
individual (\S\ref{result:fragmented}). 
In addition, differences among persona-based conditioning approaches were small by comparison and metric-dependent (\S\ref{result:persona-conditioning}).
Finally, our findings suggest population-level agreement can potentially mask individual-level error (\S\ref{result:level-of-analysis}).
Collectively, these findings indicate that no single metric is sufficient to characterize the
behavioral fidelity of a long-horizon activity simulator.

We report detailed results in \autoref{app:additional-results}, backbone-specific patterns in \autoref{app:backbone}, and a held-out evaluation in \autoref{app:holdout}.

\subsection{Statistical Priors Reduce Divergence on Activity and Sequence
Distributions}
\label{result:stat-activity-sequence}

\begin{table}[t]
    \centering
    \small
    \setlength{\tabcolsep}{3pt}
    \begin{tabular}{@{}ll cc@{}}
      \toprule
       & Method & Time allocation & Transition \\
      \midrule
      B0 & No-Persona    & $.21$ \ci{.15}{.27}             & $.36$ \ci{.29}{.46} \\
      \midrule
      C1 & Persona-Auth. & $.24$ \ci{.17}{.32}             & $.35$ \ci{.26}{.45} \\
      C2 & Persona-Inf.  & $.19$ \ci{.12}{.26}             & $.31$ \ci{.26}{.40} \\
      C3 & \,+ Few-Shot  & $.22$ \ci{.15}{.29}             & $.33$ \ci{.27}{.42} \\
      \midrule
      C4 & Stat.-Prior   & $\mathbf{.010}$ \ci{.007}{.015} & $.107$ \ci{.072}{.146} \\
      C5 & Hybrid        & $.010$ \ci{.007}{.013}          & $\mathbf{.091}$ \ci{.085}{.098} \\
      \bottomrule
    \end{tabular}
    \caption{JS divergence between simulated and real activity distributions.
  \emph{Time allocation} is the day-level activity time allocation distribution.
  \emph{Transition} is the conditional next-activity distribution.}
  \label{tab:time-transition}
\end{table}

At the day level, conditioning on statistical priors (C4--C5) reduced
divergence in activity time allocation relative to both persona-based
conditioning (C1--C3) and the no-persona control (B0), from $\text{JS} = .19$--$.24$ for
B0--C3 to $.010$ for C4--C5 (Table~\ref{tab:time-transition}).
All eight contrasts of C4 and C5 against
B0--C3 excluded zero, with the smallest, C2 vs.\ C5, at $\Delta\text{JS} = .18$
$[.09, .27]$. Activity frequency followed the same pattern,
with the smallest contrast at $\Delta\text{JS} = .11$
$[.08, .14]$ (C2 vs.\ C5).

At a local level, conditioning on statistical priors similarly reduced
divergence on transitions (Table~\ref{tab:time-transition}) and on longer sequential motifs.
Divergence in transitions fell from $\text{JS} = .31$--$.36$ (B0--C3) to $.107$ (C4) and
$.091$ (C5). All contrasts against
B0--C3 again excluded zero, with the weakest, C2 vs.\ C4, at
$\Delta\text{JS} = .21$ $[.09, .29]$. The 3-gram and 4-gram distributions
followed the same pattern, with every one of the 16 contrasts (eight per
$n$-gram order) excluding zero.

At time-of-day granularity, the priors had lower time allocation divergence in
all four windows, with all eight contrasts of C4 and C5 against B0--C3
excluding zero in every window.
%The smallest of these contrasts was C2 vs. C5 in 15:00--17:00, at $\Delta\text{JS} = .09$ $[.04, .14]$.
The time-binned activity frequency distributions showed the same pattern
throughout, with all contrasts excluding zero in every window.

\subsection{Statistical Priors Produce Fragmented, Homogeneous Days}
\label{result:fragmented}

While conditioning on statistical priors reduced divergence in activity and
sequence distributions, this fidelity advantage did not extend to other
properties of the simulated day.

First, our statistical-prior (C4) and hybrid (C5) approaches generated days
with $\Delta_{\text{sw}} = +38.0$ and $+32.3$ more activity switches than their real counterparts,
respectively (\autoref{tab:switches}).
The persona-based approaches and the control erred in the opposite direction,
producing $10.8$ to $15.0$ fewer switches across B0--C3, and remained
closer to the real count in absolute terms ($L_{1} = 13.0$--$15.2$ against $32.3$--$38.0$).
All contrasts of C4 and C5 against B0--C3 excluded zero on both the unsigned
distance and the signed bias, with the weakest at $\Delta L_{1} = -17.1$
$[-34.3, -5.2]$ (C1 vs.\ C5).

Second, every approach produced less within-person day-to-day variation than the
real data, and conditioning on statistical priors reduced it furthest (\autoref{tab:within-persona}).
Real individuals varied across their own days with a mean
pairwise session JS of $V^\text{real} = .097$.
Sessions generated under statistical priors were the most uniform, at
$V^\text{sim} = .018$ (C4) and $.026$ (C5), while the persona-based and control
conditions (B0--C3) ranged from $.043$ to $.064$
(Table~\ref{tab:within-persona}).
All contrasts of C4--C5 against B0--C3 exclude zero, except C1 vs.\ C5, at $\Delta V = .02$ $[-.01, .03]$.
The smallest contrast excluding zero was C3 vs.\ C5, at $\Delta V = .018$ $[.002, .027]$.

The mean absolute gap from real within-person variability
$L_{1}(\mathrm{JS})$ was largest for C4 and
C5 ($.079$ and $.070$, compared with $.043$--$.054$ for B0--C3).
The four contrasts between C4 and B0--C3 excluded zero, with the smallest
being its comparison with C3, at $\Delta L_{1}(\mathrm{JS}) = .02$ $[.01, .03]$.
Thus, C4 not only produced less within-person variability, but
also reproduced the observed magnitude of within-person variability less
faithfully than every persona-based approach and the no-persona control.
For C5, only its comparison with C2 excluded zero
($\Delta L_{1}(\mathrm{JS}) = .025$ $[.001, .042]$); its remaining contrasts against B0, C1,
and C3 included zero.

\begin{table}[t]
    \centering
    \small
    \setlength{\tabcolsep}{3pt}
    \begin{tabular}{@{}ll cc@{}}
      \toprule
       & Method & $L_{1}$ & $\Delta_{\text{sw}}$ \\
      \midrule
      \multicolumn{4}{@{}l}{\emph{Real days average $N_{\text{sw}}^{\text{real}} = 26.7$ activity switches} \ci{21.1}{30.4}} \\
      \midrule
      B0 & No-Persona    & $13.0$ \ci{7.6}{17.4}  & $-13.0$ \ci{-17.4}{-7.6} \\
      \midrule
      C1 & Persona-Auth. & $15.2$ \ci{10.5}{19.6} & $-10.8$ \ci{-19.1}{-4.3} \\
      C2 & Persona-Inf.  & $15.0$ \ci{8.7}{19.3}  & $-15.0$ \ci{-19.3}{-8.7} \\
      C3 & \,+ Few-Shot  & $13.4$ \ci{9.4}{17.4}  & $-13.4$ \ci{-17.4}{-9.4} \\
      \midrule
      C4 & Stat.-Prior   & $38.0$ \ci{24.3}{47.8} & $+38.0$ \ci{+24.3}{+47.8} \\
      C5 & Hybrid        & $32.3$ \ci{19.5}{45.3} & $+32.3$ \ci{+19.5}{+45.3} \\
      \bottomrule
    \end{tabular}
    \caption{The activity-switch gap between simulated and real days. $N_{\text{sw}}$ is the number of activity switches in a day.
  $\Delta_{\text{sw}}$ is the signed gap $N_{\text{sw}}^{\text{sim}} -
  N_{\text{sw}}^{\text{real}}$ and $L_{1}$ its unsigned counterpart.}
  \label{tab:switches}
  \end{table}

\begin{table}[t]
    \centering
    \small
    \setlength{\tabcolsep}{3pt}
    \begin{tabular}{@{}ll cc@{}}
      \toprule
       & Method & $V^{\text{sim}}$ & $L_{1}(\mathrm{JS})$ \\
      \midrule
      B0 & No-Persona    & $.058$ \ci{.039}{.066} & $.043$ \ci{.027}{.060} \\
      \midrule
      C1 & Persona-Auth. & $.043$ \ci{.036}{.053} & $.054$ \ci{.038}{.066} \\
      C2 & Persona-Inf.  & $.064$ \ci{.061}{.069} & $.045$ \ci{.026}{.064} \\
      C3 & \,+ Few-Shot  & $.044$ \ci{.036}{.050} & $.054$ \ci{.037}{.088} \\
      \midrule
      C4 & Stat.-Prior   & $.018$ \ci{.014}{.027} & $.079$ \ci{.051}{.104} \\
      C5 & Hybrid        & $.026$ \ci{.019}{.043} & $.070$ \ci{.037}{.093} \\
      \midrule
      \multicolumn{2}{@{}l}{Real} & $.097$ \ci{.077}{.119} & --- \\
      \bottomrule
    \end{tabular}
    \caption{Within-person session-to-session variability. $V^{\text{sim}}$ is the mean pairwise JS between sessions for the same individual; $L_{1}(\text{JS}) = |V^{\text{sim}} - V^{\text{real}}|$ is the gap to the real reference.}
  \label{tab:within-persona}
  \end{table}

\subsection{Persona-Based Conditioning Differences Are Limited and Metric-Dependent}
\label{result:persona-conditioning}

Across our metrics, the persona-based conditioning approaches and the no-persona control
generally differed less than the statistical-prior approaches did against them.
On day-level activity time allocation, for example, the six pairwise contrasts
among B0--C3 span $|\Delta\text{JS}| = .008$--$.056$, whereas the eight
contrasts of C4 and C5 against B0--C3 span $|\Delta\text{JS}| = .179$--$.235$ (\autoref{tab:time-transition}).

We computed all six pairwise contrasts among B0--C3 for each of our $13$
distributional metrics, giving $78$ contrasts in total, of which $21$ excluded
zero.
Within these, trace-inferred persona descriptors (C2) had lower divergence
than every other approach in $38$ of the $39$ contrasts involving it
($3$ comparisons $\times$ $13$ metrics); $13$ of those $39$ excluded zero.
The remaining approaches showed no comparable trend, and their contrasts
pointed in different directions depending on the metric.
For example, of the five contrasts between B0 and C1 that excluded zero, three favored C1, while two favored B0. 
C1 and C3 divided similarly: of their three contrasts that excluded zero, two
favored C3 and one favored C1.
No contrast between B0 and C3 excluded zero on any metric.

These results suggest that the choice among persona-based conditioning approaches had a small effect on 
behavioral fidelity relative to the difference between persona-based conditioning and statistical priors. 
Within that smaller range, trace-inferred descriptors (C2) showed lower
divergence in all but one of its contrasts, but only a third of them excluded
zero.
For the remaining approaches, whether one approach improved fidelity over another depended on the metric used to assess it.

\subsection{Behavioral Fidelity May Depend on the Level of Analysis}
\label{result:level-of-analysis}

Finally, the values reported
above are computed on an individual basis. Computing a population-averaged
distribution first and then measuring its divergence from the real data
systematically yields smaller values.
Contrasting the individual- and population-level JS of each of our $13$
distributional metrics, all $13$ excluded
zero, including activity time allocation
($\Delta_{\text{ind}-\text{pop}} = +.020$ $[+.011, +.023]$) and
transition ($+.069$ $[+.040, +.085]$) distributions.
We take this to suggest that reporting only
population-level agreement
may understate simulator error on individual routines. 
Specifically,
a simulator may produce a realistic \emph{average} day across the population
without maintaining fidelity for individuals.

\section{Conclusion}
\label{sec:conclusion}
In this work, we contribute an evaluation framework for the behavioral fidelity
of long-horizon human simulation. As a case study, we evaluated six
simulation approaches on a newly collected dataset of human activities captured within an
office. 
Our results show that, in this experimental setting, while methods
leveraging statistical priors reduce divergence on activity distributions and
local transition structure, they over-segment the day relative to real
individuals and produce routines that vary far less from day to day than real
ones.
In addition, our results suggest that differences among persona-based conditioning approaches are small by comparison and depend on the metric used to assess them.
Finally, we find that population-level
agreement can mask individual-level error. 
Together, these results highlight the need for a holistic approach to evaluating LLM-based agents as human proxies.

\section{Limitations}
\label{sec:limitations}

\paragraph{Generalizability.}
Our framework is setting-agnostic. 
Its metrics apply to arbitrary categorical activity sequences and assume no particular label vocabulary, physical layout, or social configuration. 
Our experiments are a case study showing why evaluation at a single granularity, or with a single aggregate metric, can be insufficient.
Our results, however, are specific to our experimental setting (single multi-room office, one work week, five individuals, six activity labels, fixed social configuration), and need replication across environments, longer observation periods, richer vocabularies, and varied social configurations.
Such evaluation requires an uncommon form of ground truth: unscripted, multi-day, person-specific sequences with temporally complete labels per session. 
Few resources meet this requirement (\autoref{app:additional-datasets}), and we see value in expanding them. 

\paragraph{Behavior without Intent.}
Our annotations do not capture mental state, so our framework evaluates the fidelity of
\emph{outward} activity sequences only.
This is necessary but not sufficient for
downstream uses that depend on intent. 
It also limits counterfactual inference. 
Behavioral traces alone do not identify why a pattern holds, so a simulator matched to observed sequences may not reproduce behavior under changed conditions. 
Future work could pair traces with self-reported intent from experience sampling or interviews.

\paragraph{Simulator design.}
Several agent design choices, while informed by related agent
architectures, warrant further investigation. Our agent selects activities
partly by reference to a day plan it generates (\S\ref{sec:planning}), but we did not systematically vary plan
quality, so how much the plan shapes generation outcomes remains
unvalidated.
Our statistical priors likewise condition only on first-order transitions. Higher-order priors may yield greater behavioral coherence.
Moreover, persona-based prompting yielded modest and metric-dependent effects here, but whether this holds for long-horizon simulations in general remains an interesting open question.
Finally, while our results were generally upheld in the three models we ran, they notably shifted in magnitude (\autoref{app:backbone}). 
How backbone choice interacts with conditioning approach is, therefore, an open question.
\section{Ethics Statement}
\label{sec:ethics}
\paragraph{Consent and Data Release.}
Our data collection was approved by our
Institutional Review Board (IRB STUDY2025\_00000414). 
Occupants and visitors were informed of continuous multi-camera recording through signs at the entrance and throughout the workspace and through direct messages in its shared communication channel, both before and during data collection.
These notices also stated their right to request removal of their data at any time; no such requests were received.

To protect privacy, we release only derived activity
sequences, consisting of per-frame activity labels, room labels, and estimated
2D positions on a scaled floor plan, each associated with an anonymized individual
identifier (\autoref{app:dataset}).
The released dataset does not contain raw video or identity-linkable metadata.
Trajectory data can still carry re-identification risk for
individuals with knowledge of the environment. 
To mitigate this, sessions are released without calendar dates, with timestamps retaining only the time of day.
The activity vocabulary is also coarse, limiting the behavioral detail attributable to any individual.

\paragraph{Risks of Workplace Surveillance.}
Instrumenting an office with cameras also raises power asymmetries: visibility varies by room, and visitors and transient
occupants have less capacity to opt out than regular occupants. 
This is part of
why we release only derived data and withhold all imagery. 
We ask that researchers treat this resource as sensitive workplace trace data, not as precedent for monitoring where occupants have not been informed and given the opportunity to opt out.

\paragraph{Misuse of Behavior Simulation.}
By making the gap between
simulated and real behavior \emph{measurable}, we hope to support progress toward more faithful long-horizon simulation. 
Improved fidelity should
not, however, be taken as license to substitute simulated agents for people,
particularly in high-stakes decisions.
We see such simulations as a complementary probe for studying human
behavior~\cite{li2026whatif}, and their application should be accompanied by independent ethical review
and empirical grounding.
\section*{Acknowledgments}
We thank Sherry Tongshuang Wu, Jacob Springer, Leena Mathur, Drishti Goel, Yi-Hao Peng, Alexander Wang, and Jarod Bloch for their feedback and support throughout this project. This work was funded in part by Fujitsu Ltd.

\bibliography{references}

\appendix

\section{Behavioral Fidelity Metrics}
\label{app:metrics}

In this section, we give operational definitions for the behavioral metrics
introduced in \S\ref{sec:framework} and summarized in
Table~\ref{tab:taxonomy}.

As defined in \S\ref{sec:problem},
we let $\sigma = (a, t_{\text{start}}, t_{\text{end}})$ denote an \emph{activity
episode}, an uninterrupted stretch of a single activity, labeled $a$ from a
closed activity vocabulary $\mathcal{A}$, running from time $t_{\text{start}}$
to time $t_{\text{end}}$.
We represent behavior as an \emph{activity trajectory}
$\tau = (\sigma_1, \ldots, \sigma_M)$, where $M$ is the number of episodes it
contains.

\paragraph{Behavioral descriptors.}
For a trajectory $\tau$, we can first compute the following descriptors with respect to its activities. 
\begin{itemize}

\item \textbf{Episode duration.} The average duration of a single episode of
  each activity $a \in \mathcal{A}$ in $\tau$.

\item \textbf{Transitions.} The distribution over the next activity given the
  current one, estimated from the adjacent episode pairs
  $(\sigma_{m-1}, \sigma_m)$ in $\tau$.

\item \textbf{Behavioral motifs.} The distribution over short activity
  sequences, taken as the 3- and 4-grams formed by consecutive episodes in
  $\tau$.

\item \textbf{Time allocation.} The proportion of $\tau$'s total duration spent
  in each activity $a \in \mathcal{A}$.

\item \textbf{Frequency.} The proportion of $\tau$'s $M$ episodes labeled with
  each activity $a \in \mathcal{A}$.

\item \textbf{Switches.} The number of times the activity changes across $\tau$,
  equal to $M - 1$ since consecutive episodes carry different labels.
\end{itemize}

The descriptors above characterize a single trajectory.
Given the set of trajectories belonging to one individual,
$\{\tau_i^{(s)} : s \in \mathcal{S}_i\}$, we can further characterize that
individual's \textbf{within-person variability}: how much their behavior differs between sessions.
We compute a behavioral descriptor for each of their trajectories and take the
mean difference over all pairs.
Here, we characterize within-person variability with respect to the time
allocation distribution, using the mean pairwise JS divergence between an
individual's trajectories.

Given trajectories belonging to multiple individuals, we can likewise
characterize a population's \textbf{between-person variability}: how much
individuals differ from one another.
We compute a descriptor for each trajectory, average these within each
individual, and take the mean difference over all pairs of individuals.
Again we characterize this with respect to the time allocation distribution,
using the mean pairwise JS divergence between individuals' descriptors.

Each of these descriptors can be computed over the full trajectory or over a
time window $[u, v)$ within it, in which case we first restrict $\tau$ to the
window, clipping each episode to its intersection with $[u, v)$.
This yields the time-of-day descriptors of Table~\ref{tab:taxonomy}.

\paragraph{Quantifying differences.}
Given two trajectories, we quantify the difference between them with respect to
each of the behavioral descriptors above.

For the distributional descriptors, namely transitions, behavioral motifs, time
allocation, and frequency, we compute
total variation distance (TV),
Jensen--Shannon divergence (JS),
and the forward and backward
Kullback--Leibler divergences (KL).

For the remaining descriptors, namely episode duration and switches,
since they are not probability distributions,
we instead compute an unsigned distance ($L_{1}$) and a signed bias ($\Delta$).
Within- and between-person variability are likewise not distributions but single
summary values, and for these we report the unsigned distance ($L_{1}$) between
the simulated and real values.

\paragraph{Characterizing Fidelity.}

Now, given a simulated set of trajectories
$\mathcal{D}_{\text{sim}} = \{\tilde{\tau}_i^{(s)} : i \in \mathcal{I},\, s \in
\tilde{\mathcal{S}}_i\}$ and a reference set of trajectories
$\mathcal{D}_{\text{real}} = \{\tau_i^{(s)} : i \in \mathcal{I},\, s \in
\mathcal{S}_i\}$, we can quantitatively characterize the fidelity of the simulator.

Fidelity is first characterized at the \textbf{individual level}, that is,
whether the simulator faithfully reproduces the behavior of each individual.
For an individual $i \in \mathcal{I}$, we compute a behavioral descriptor for
each of their real and simulated trajectories, and aggregate these into a single
real and a single simulated descriptor for that individual by averaging across
trajectories.
We then quantify the difference between the two aggregates as above.
Averaging this difference across all represented individuals yields a single
value per metric for the simulator.

Alternatively, fidelity can be characterized at the \textbf{population level},
that is, whether the simulator faithfully reproduces the behavior of the
population as a whole.
For this, we compute a single real and a single simulated descriptor for the
population, averaging each individual's trajectory descriptors as before and
then averaging across individuals.
Quantifying the difference between these two aggregates as above yields a single
value per metric.

\section{Data Release}
\label{app:dataset}
We release the derived activity sequences detailed in \S\ref{sec:data}.
Raw video is withheld to protect participant privacy.
Each session covers one day of activity
and is distributed as a sequence of frames sampled at 1-minute intervals.
Each frame records its timestamp, the number of people present, and a per-individual
record containing an anonymized individual ID, an activity label, a room label, and
an estimated 2D position on a scaled floor plan.
Per-individual records labeled \emph{engaging in conversation} additionally carry a
conversation ID.
The scaled floor plan is included with the release.

While our own experiments use only the activity sequences, we release the
positional and conversational data as well, since they may support work on
other dimensions of behavior, such as spatial routine, co-presence, and social
interaction. 
We note that only activity labels were multiply annotated.
Positions, room labels, and conversation IDs were
manually verified by the first author, but not independently validated.

\section{Dataset Construction}
\label{app:construction}
The dataset was constructed in three stages: tracking and localization (\ref{app:tracking}), activity
annotation (\ref{app:annotation}), and conversation labeling (\ref{app:conversation}).

\subsection{Tracking and Localization}
\label{app:tracking}
This stage processes the raw multi-view video into persistent, globally
identified trajectories within the environment. 
It consists of local single-camera tracking, multi-camera spatial fusion, and cross-session identity stitching. 
We additionally assign per-frame room labels and manually inspect
and correct the pipeline outputs.

\paragraph{Single-camera Tracking.}
Each video feed is first processed individually. 
We use YOLO26L-Pose~\cite{jocher2026ultralyticsyolo26} to identify individuals in the environment.
For each detected individual, the system extracts bounding boxes and skeletal keypoints on a frame-by-frame basis.
To establish temporal consistency within a single viewpoint, these frame-level spatial detections are linked using a SAM3-based tracking module~\cite{carion2026sam3}.
This generates continuous, reliable tracklets for each individual within a camera's field of view.

\paragraph{Multi-camera Fusion.}
All cameras are pre-calibrated, and we project single-camera tracklets onto a
shared 2D floor plan by mapping camera-view pixels to global physical
coordinates via homography. Because an individual is often captured simultaneously by
multiple overlapping cameras, we then resolve redundant tracks by clustering
projected tracks on a combination of spatial proximity on the floor plan and
visual appearance similarity, fusing single-view tracklets into environment-wide
identities.

\paragraph{Global Identification.}
This final step bridges discontinuous tracking sessions, \eg when an individual
leaves the environment and returns later.
For each track we compute a global
appearance representation by aggregating and normalizing its visual features
across all frames, and assign cross-session identities by agglomerative
average-link clustering on the cosine similarity between these representations.
To ensure physical plausibility, the clustering enforces a hard temporal
constraint: any two tracks with overlapping temporal intervals are blocked from
merging, so a single identity cannot exist in two places at once.

\paragraph{Room Labels.}
Each individual is assigned a room label at each frame based on camera visibility.
Because each camera view largely captures a single room, the set of cameras in
which an individual is detected is informative of their location.
We assign an initial room label by majority vote over the rooms associated with the cameras that detect the individual in that frame.
We found this more robust than testing projected positions against room boundaries, given projection errors.
Ambiguous cases, including ties, are flagged for manual resolution.

\paragraph{Manual Refinement.}
Using a custom tool that allows switching between synchronized camera views, the first author manually inspected all recordings in full, correcting
mis-identifications and resolving flagged room-label ambiguities.

\subsection{Activity Annotation}
\label{app:annotation}
This stage assigns a frame-level activity label to each tracked individual.
We first develop an activity vocabulary, then recruit and train three annotators,
establish inter-rater reliability on a shared sample, and annotate the remaining
frames.

\paragraph{Activity vocabulary.}
The first author developed the activity vocabulary following
\citet{patidar2025organichar}. The author iteratively open-coded a subset of
the sequences, grouped the resulting codes into candidate categories, and merged
overlapping categories until the label set stabilized, yielding six activity
labels (\S\ref{sec:activity-vocabulary}).

\paragraph{Annotator training.}
We recruited three annotators from our university, compensated at $\$20$ per hour, and trained them in person.
Following an overview of the project, they were introduced to our
custom annotation tool, which loads individual sessions, allows switching
between synchronized camera views, and supports frame-level activity and
location labels for each individual. The annotators then completed a two-hour joint
calibration session with the first author, annotating footage together and
discussing disagreements until reaching consensus on the use
of each label.

\paragraph{Inter-rater reliability.}
The three annotators and the first author then independently labeled the same
200 frames, sampled in a stratified manner across sessions and time.
Agreement on activity labels was high (Fleiss' $\kappa=0.86$): all four raters assigned the same label on $87\%$ of timestamps (mean pairwise agreement $92.7\%$). 

\paragraph{Full activity annotation.}
Given this level of agreement, the remaining frames were single-coded. Each annotator independently annotated one full day of activity and the first author
annotated two, yielding five sessions in total.

\subsection{Conversation Labeling}
\label{app:conversation}
Finally, for individual records labeled \emph{engaging in conversation}, the first
author assigned a shared conversation ID to individuals engaged in a common
interaction and tracked that ID across frames as individuals joined and left.

\section{Backbone Analysis}
\label{app:backbone}

The results presented in \S\ref{sec:results} are computed as an average
over three LLM backbones: Llama~3.3~70B~Instruct, GPT-5.4-mini, and
Gemini~2.5~Flash.
Here, we report an exploratory, descriptive breakdown of those results
by backbone.

\begin{table}[t]
  \centering
  \small
  \setlength{\tabcolsep}{4pt}
  \begin{tabular}{@{}ll cccc@{}}
    \toprule
     & Method & Gemini & GPT & Llama & Range \\
    \midrule
    B0 & No-Persona    & $.206$ & $.228$ & $.207$ & $.022$ \\
    \midrule
    C1 & Persona-Auth. & $.271$ & $.259$ & $.205$ & $.067$ \\
    C2 & Persona-Inf.  & $.189$ & $.214$ & $.164$ & $.050$ \\
    C3 & \,+ Few-Shot  & $.208$ & $.244$ & $.213$ & $.035$ \\
    \multicolumn{2}{@{}l}{Persona methods (mean)}
                               &        &        &        & $.044$ \\
    \midrule
    C4 & Stat.-Prior   & $.013$ & $.008$ & $.008$ & $.004$ \\
    C5 & Hybrid        & $.014$ & $.006$ & $.011$ & $.008$ \\
    \multicolumn{2}{@{}l}{Statistical priors (mean)}
                               &        &        &        & $.006$ \\
    \bottomrule
  \end{tabular}
  \caption{JS divergence between simulated and real day-level activity time-allocation distributions. 
  \emph{Range} is the difference between the largest and smallest of
  the three backbone scores, and the mean rows average those ranges
  over B0--C3 and over C4--C5.}
  \label{tab:backbone-day-js}
\end{table}

First, our analysis suggests that the differences between the behaviors
generated with the statistical priors (C4--C5) and those generated with
the control and the persona-based approaches (B0--C3) hold across
backbones.
Taking each backbone separately, the statistical priors reduced
divergence in day-level activity time-allocation
(\autoref{tab:backbone-day-js}), transitions, and longer sequential
motifs relative to persona conditioning, as well as in time-allocation
and activity frequency within each of the four two-hour windows;
however, they also generated days with more activity switches than real
ones and with less day-to-day variation within an individual.
Backbone choice seemingly changed magnitudes, but not these qualitative
conclusions.

Second, we observe that the backbone generally had a larger effect on
the persona-based approaches on activity and transition distributions.
For day-level activity time-allocation (\autoref{tab:backbone-day-js}),
the cross-backbone JS spread was comparatively larger for the 
persona-based approaches (B0--C3: $.044$) than for the statistical
priors (C4--C5: $.006$).
The corresponding ranges were $.040$ against $.003$ for activity
frequency and $.063$ against $.026$ for transitions.
For time-allocation and activity frequency, the same patterns
replicated across time windows.
% This spread appeared to be comparable to the variation across the
% persona-based approaches themselves, but small compared to the
% difference between those approaches and the statistical priors.

Our results do not allow us to determine why these cross-backbone
differences occurred and thus do not read them as a finding about
particular models.
How the backbone model, including factors like pretraining data
composition, model scale, and post-training, affects simulation
fidelity requires further investigation.

\section{Leave-One-Session-Out Evaluation}
\label{app:holdout}

In \S\ref{sec:experiments}, we report on results from evaluating simulators
whose conditioning was derived from the full real activity corpus, measuring
each approach's ability to replicate observed behavior. 
To assess generalization across sessions, we conducted a preliminary
leave-one-session-out evaluation.

\paragraph{Setup.}
We hold out one session at a time, excluding it from trace-derived
components of the conditioning, namely the trace-inferred behavioral
tendencies (C2), few-shot sequences (C3), and statistical priors (C4--C5). B0 and C1 use no trace-derived conditioning and are unchanged across folds. 

We use 
$\mathcal{D}_{\text{real}}$ from \S\ref{sec:data}, which comprises five
recorded sessions, giving five folds and $5 \times 6 = 30$ simulations. 
Due to cost constraints, we conducted experiments using a single backbone,
GPT-5.4-mini, and one generation per fold. 
We add a data-only reference, DATA.
For each fold, this calculates the mean distribution of an individual's four
non-held-out sessions. DATA is not a simulator; rather, it provides an
empirical reference for how well behavior from the observed sessions transfers to the held-out session.

\begin{table*}[t]
  \centering
  \small
  \setlength{\tabcolsep}{3pt}
  \begin{tabular}{@{}l cc cc cc@{}}
    \toprule
    Method & Time allocation & Transition & \multicolumn{2}{c}{Activity-switch gap} & \multicolumn{2}{c}{Session variability} \\
    \cmidrule(lr){2-2} \cmidrule(lr){3-3} \cmidrule(lr){4-5} \cmidrule(lr){6-7}
    & JS & JS & $L_{1}$ & $\Delta_{\text{sw}}$ & $V^{\text{sim}}$ & $L_{1}(\mathrm{JS})$ \\
    \midrule
    B0 No-Persona    & $.30$ \ci{.21}{.36}             & $.36$ \ci{.33}{.39}          & $16.8$ \ci{11.2}{21.2} & $-16.6$ \ci{-21.1}{-10.7} & $.038$ \ci{.026}{.048} & $.058$ \ci{.037}{.092} \\
    \midrule
    C1 Persona-Auth. & $.28$ \ci{.21}{.37}             & $.39$ \ci{.38}{.42}          & $18.8$ \ci{14.2}{24.1} & $-18.8$ \ci{-24.1}{-14.2} & $.041$ \ci{.032}{.054} & $.056$ \ci{.024}{.076} \\
    C2 Persona-Inf.  & $.30$ \ci{.22}{.39}             & $.50$ \ci{.41}{.56}          & $19.6$ \ci{14.3}{23.2} & $-19.6$ \ci{-23.2}{-14.3} & $.052$ \ci{.041}{.070} & $.051$ \ci{.031}{.075} \\
    C3 \,+ Few-Shot  & $.29$ \ci{.21}{.38}             & $.41$ \ci{.33}{.51}          & $17.5$ \ci{13.7}{21.0} & $-17.5$ \ci{-21.0}{-13.7} & $.059$ \ci{.043}{.085} & $.055$ \ci{.046}{.073} \\
    \midrule
    C4 Stat.-Prior   & $\mathbf{.071}$ \ci{.066}{.078} & $.35$ \ci{.29}{.41}          & $32.5$ \ci{22.2}{41.2} & $+32.5$ \ci{+22.2}{+41.2} & $.022$ \ci{.013}{.032} & $.074$ \ci{.043}{.101} \\
    C5 Hybrid        & $.075$ \ci{.055}{.096}          & $\mathbf{.32}$ \ci{.27}{.39} & $27.2$ \ci{22.6}{38.0} & $+27.2$ \ci{+22.6}{+38.0} & $.023$ \ci{.015}{.028} & $.073$ \ci{.053}{.093} \\
    \midrule
    DATA             & $.065$ \ci{.050}{.087}          & $.26$ \ci{.23}{.28}          & $5.8$ \ci{3.7}{8.3}    & ---$^{\dagger}$           & ---$^{\dagger}$        & ---$^{\dagger}$        \\
    \midrule
    Real             & ---                             & ---                          & ---                    & ---                       & $.097$ \ci{.076}{.119} & ---                    \\
    \bottomrule
  \end{tabular}
  \caption{Leave-one-session-out results. \emph{Time allocation} JS is the
  day-level activity time-allocation divergence and \emph{Transition} JS the
  conditional next-activity divergence to the held-out session.
  $L_{1}$ and
  $\Delta_{\text{sw}}$ are the unsigned and signed gaps in activity switches
  against real days. $V^{\text{sim}}$ is the mean pairwise JS between an
  individual's simulated sessions and $L_{1}(\mathrm{JS})$ its absolute difference
  from the same quantity over their real sessions. DATA is the fold-wise mean
  distribution over the four non-held-out sessions. $^{\dagger}$Not reported
  for DATA: averaging leave-one-out estimates recovers the full-sample value by
  construction, so DATA's signed switch gap is necessarily zero and its session
  variability necessarily $V^{\text{real}}$.}
  \label{tab:holdout}
\end{table*}

\paragraph{Metrics.}
For most of the metrics we evaluate (e.g., activity time allocation,
frequency, switches), we first calculate the divergence on a per-fold basis. In
other words, for each fold, we only calculate divergence between the metric
computed over the simulated activity sequence and the real held-out session.
These fold values are then averaged within an individual, and finally across
individuals. One exception is within-person variability, which quantifies
differences within an individual's simulated sessions. For this, we compute the
mean pairwise divergence among an individual's simulated sessions across all
folds ($V^{\text{sim}}$), and its absolute difference from the same quantity
computed over their real sessions ($L_1(\mathrm{JS})$). These per-individual
values are averaged across individuals.
We compute 95\% BCa bootstrap confidence intervals ($B = 10{,}000$) 
for per-approach metrics and paired contrasts by resampling over individuals with replacement, using each individual's full set of folds; 
contrasts are Bonferroni-corrected across the $\binom{6}{2} = 15$ simulator pairs.
DATA is reported as a descriptive reference and does not enter any contrast.
Due to the small number of individuals ($n=5$), 
these intervals characterize how much each estimate depends on which individuals are included, rather than supporting generalization beyond this cohort.

\paragraph{Results.}

Directionally, the hold-out experiments support the results in
\S\ref{sec:results}.
The results showed that statistical priors reduced divergence in activity and
sequence distributions.
For activity time allocation, the statistical priors (C4--C5) reported a
divergence of $.071$--$.075$, whereas the persona-based approaches and the
no-persona control (B0--C3) reported $.28$--$.30$.
All eight contrasts of C4--C5 against B0--C3
exclude zero, the weakest being C1 vs.\ C5 at $\Delta\text{JS} = .209$
\ci{.106}{.370}.
Activity frequency, 3-grams, and 4-grams follow the same pattern, with all
eight cross-group contrasts excluding zero on each.
Transitions are the exception.
Here the priors reported $.32$--$.35$ against $.36$--$.50$ for B0--C3, and
only two contrasts exclude zero, C2 vs.\ C4 at $.159$
\ci{.051}{.299} and C2 vs.\ C5 at $.187$ \ci{.068}{.272}.

Our results similarly showed statistical priors generating more fragmented and
homogeneous days.
For the activity-switch gap, C4 and C5 produced $32.5$ and $27.2$ more
switches than their held-out counterparts, while B0--C3 produced $16.6$--$19.6$
fewer.
All eight cross-group contrasts exclude zero on
the signed bias; the unsigned distance followed the same pattern, but three of
the eight included zero.
For within-person variability, sessions generated under the priors were again
the most uniform ($V^{\text{sim}} = .022$ for C4 and $.023$ for C5, against
$.038$--$.059$ for B0--C3 and $V^{\text{real}} = .097$).
Four of the eight cross-group contrasts on
$V^{\text{sim}}$ exclude zero, the weakest being C2 vs.\ C5 at $.028$
\ci{.009}{.056}.
For the gap to real variability, C4 and C5 were furthest from the real value
($L_{1}(\mathrm{JS}) = .074$ and $.073$, against $.051$--$.058$ for B0--C3),
though only two contrasts exclude zero, both against C2: C4 at
$.023$ \ci{.007}{.043} and C5 at $.022$ \ci{.009}{.047}.

Differences among the persona-based approaches were again limited.
On day-level activity time allocation, the six pairwise contrasts among B0--C3
span $|\Delta\text{JS}| = .005$--$.020$, against $.209$--$.234$ for the eight
contrasts of C4--C5 against B0--C3.
Computing the same six contrasts for each of the $13$ distributional metrics
used in \S\ref{result:persona-conditioning} gives $78$ contrasts, of which
$12$ exclude zero, against $21$ in the main study.
Six of the $30$ contrasts across the five quantities in \autoref{tab:holdout}
likewise exclude zero.
The consistent advantage for trace-inferred descriptors (C2) observed in
\S\ref{result:persona-conditioning} does not survive here; instead, C2 becomes
metric-dependent in the same way the other approaches were.
C2 has higher divergence in $31$ of the $39$ contrasts involving it and lower
divergence in the remaining eight.
Eight of the $39$ exclude zero, all of them contrasts on which C2 is higher.
Similarly, other pairs did not separate consistently.
B0, for example, has lower divergence than C1 in six of the $13$ metrics and
higher in the other seven, with one contrast excluding zero.

Population-level divergence is lower than individual-level divergence in $76$
of the $78$ method--metric combinations we evaluate ($6$ simulators $\times$
$13$ distributional metrics).
For C4--C5, transition divergence falls from
$.331$ at the individual level to $.198$ at the population level.

The DATA reference provides useful context. Statistical priors approach it on
activity time allocation ($.071$ and $.075$ against $.065$) and on transitions
($.35$ and $.32$ against $.26$), but every simulator is far off on activity
switches, ranging from $16.8$ for B0 to $32.5$ for C4 against DATA's $5.8$.
Against the real reference, every simulator again falls
well short on variability ($.022$--$.059$ against $V^{\text{real}} = .097$).

These findings should be interpreted cautiously.
The evaluation uses a single
backbone and a single generation per fold.
Moreover, the corpus contains
relatively routine behavior across only five sessions, which may make held-out
sessions comparatively predictable based on the remaining sessions.
Finally, this experiment evaluates new sessions from the same individuals, not unseen individuals. Cross-individual and cross-setting generalization remain important directions for future work (\S\ref{sec:limitations}).

\section{Datasets for Evaluating Long-Horizon Simulations}
\label{app:additional-datasets}
 
\begin{table}[t]
\centering
\footnotesize
\setlength{\tabcolsep}{3.5pt}
\begin{tabular}{@{}llcccc@{}}
\toprule
Dataset & Setting & Unscr. & Days & Seq. & $n$ \\
\midrule
Ego4D          & varied         & $\checkmark$ & $\times$ & $\times$     & 1$^{\dagger}$ \\
EgoLife        & shared house   & $\sim$       & 7        & $\times$     & 6 \\
CASTLE2024     & vacation home  & $\sim$       & 4        & $\times$     & 10 \\
ExtraSensory   & daily life     & $\checkmark$ & 7.6      & $\times$     & 1 \\
ATUS           & daily life     & recall       & 1        & $\checkmark$ & 1 \\
UKTUS          & daily life     & diary        & 2        & $\checkmark$ & hh \\
CASAS Aruba    & private home   & $\checkmark$ & $\approx$220 & $\checkmark$ & 1 \\
ARAS           & private homes  & $\checkmark$ & 30       & $\checkmark$ & 2 \\
\midrule
Ours           & shared office  & $\checkmark$ & $4\pm1$  & $\checkmark$ & 5 \\
\bottomrule
\end{tabular}
\caption{Resources for long-horizon behavioral evaluation.
\emph{Unscr.}: unscripted in-the-wild behavior ($\sim$: organized co-living event with suggested or assigned activities; \emph{recall}: day-before recall interview; \emph{diary}: in situ self-completion diary).
\emph{Days}: observed days per person ($\times$: not released as multi-day per-person records).
\emph{Seq.}: a temporally complete categorical activity sequence per person for each observed session.
$n$: number of co-located, individually labeled people (\emph{hh}: all household members).
$\dagger$: Ego4D includes 224 hours of synchronized capture from multiple co-located camera wearers, but without parallel per-person activity sequences.}
\label{tab:datasets}
\end{table}
 
Our analysis requires an uncommon form of ground truth. For each person, we need an activity sequence that covers an extended observation session, ideally across multiple days of unscripted behavior. 
In social settings, each co-located person must be labeled separately.

Below, we document the resources we identified as most relevant (Table~\ref{tab:datasets}). Ego4D \cite{grauman2022cvpr} records unscripted behavior in the wild, but annotates short sequences rather than continuous, day-long sessions for the same individuals. EgoLife \cite{yang2025egolife} and CASTLE2024 \cite{rossetto2025castle} record several consecutive days, but both were collected through organized co-living events, whereas our individuals followed their ordinary work routines. Their released annotations are also not activity sequences. ExtraSensory \cite{vaizman2018extrasensory} observes about a week per user, annotating each minute with a set of context labels. ATUS \cite{bls2026atus,flood2025atus} records one day per respondent as a sequence of timed activity episodes, recalled through an interview. UKTUS \citep{gershuny2017uktus} records two such days, one weekday and one weekend, as self-completed diaries in 10-minute slots. ExtraSensory, ATUS, and UKTUS, notably, do not describe the physical environment, whereas ours is represented as a scene graph. This does not preclude use of our framework, but it does not support a direct extension of our agent evaluation. The closest resources are the annotated free-living homes of CASAS \citep{cook2013casas} and ARAS \citep{alemdar2013aras}. These record residents in their own homes, observed for a month or more. 
To our knowledge, ARAS and CASAS are the most immediately usable targets for testing whether the observed results generalize.
\section{Additional Results}
\label{app:additional-results}

This appendix reports the per-approach estimates and the paired contrasts
behind the claims in \S\ref{sec:results}. Throughout, we report
JS divergence for distributional metrics and $L_1$ with a
signed bias $\Delta$ for scalar metrics, computed at the individual level and averaged over the three LLM backbones. 
Total variation (TV), both directions of KL divergence, the population-level estimates, the full per-backbone grid, and every pairwise contrast are released as a CSV alongside the dataset.
Throughout, approaches are abbreviated as in Table~\ref{tab:variants}:
\textbf{B0}~\textsc{None} (no-persona control),
\textbf{C1}~\textsc{Authored Persona},
\textbf{C2}~\textsc{Inferred Persona},
\textbf{C3}~\textsc{Persona + Few-Shot},
\textbf{C4}~\textsc{Statistical Prior}, and
\textbf{C5}~\textsc{Hybrid Prior} (\S\ref{sec:conditioning}).

\paragraph{Per-approach Estimates.}
% \label{app:results-per-approach}
\autoref{tab:app-distributional} reports all 13 distributional metrics for
each of the six approaches, and \autoref{tab:app-scalar} the scalar ones.

% Originally generated by analysis/09_render_appendix_tables.py, then edited by
% hand (time-of-day labels). Regenerating will revert those edits.
% Source: analysis/confidence_intervals_absolute_10000/absolute_metrics_by_method.csv
% Bootstrap: confidence_intervals_absolute_10000 (per-approach), confidence_intervals_v2_10000 (contrasts).
% Method mapping: none = B0, baseline = C1, extracted = C2,
%                 few_shot = C3, markov = C4, hybrid = C5.
%
% Rows are the 13 distributional metrics referenced in Section 7; columns
% are the six conditioning approaches. Lower is closer to real behaviour;
% the best approach in each row is bold.

\providecommand{\ci}[2]{{\scriptsize\,[#1,\,#2]}}

\begin{table*}[t]
  \centering
  \footnotesize
  \setlength{\tabcolsep}{2.5pt}
  \begin{tabular}{lcccccc}
    \toprule
    Metric & B0 & C1 & C2 & C3 & C4 & C5 \\
    \midrule
    Time-allocation          & $.21$ \ci{.15}{.27} & $.24$ \ci{.17}{.32} & $.19$ \ci{.12}{.26} & $.22$ \ci{.15}{.29} & $\mathbf{.010}$ \ci{.007}{.015} & $.010$ \ci{.007}{.013} \\
    Frequency                & $.17$ \ci{.13}{.20} & $.18$ \ci{.15}{.22} & $.14$ \ci{.11}{.17} & $.17$ \ci{.13}{.20} & $\mathbf{.017}$ \ci{.014}{.020} & $.026$ \ci{.018}{.034} \\
    \midrule
    Time-allocation, 09:00--11:00  & $.44$ \ci{.24}{.65} & $.42$ \ci{.23}{.63} & $.41$ \ci{.22}{.64} & $.44$ \ci{.24}{.66} & $\mathbf{.022}$ \ci{.018}{.031} & $.038$ \ci{.021}{.066} \\
    Time-allocation, 11:00--13:00  & $.37$ \ci{.27}{.47} & $.40$ \ci{.28}{.54} & $.30$ \ci{.20}{.40} & $.34$ \ci{.24}{.47} & $\mathbf{.030}$ \ci{.017}{.056} & $.031$ \ci{.025}{.045} \\
    Time-allocation, 13:00--15:00  & $.17$ \ci{.14}{.22} & $.16$ \ci{.10}{.19} & $.14$ \ci{.10}{.21} & $.17$ \ci{.14}{.22} & $.050$ \ci{.030}{.097}          & $\mathbf{.042}$ \ci{.025}{.060} \\
    Time-allocation, 15:00--17:00  & $.15$ \ci{.11}{.20} & $.20$ \ci{.16}{.27} & $.13$ \ci{.10}{.16} & $.16$ \ci{.11}{.20} & $\mathbf{.028}$ \ci{.014}{.046} & $.041$ \ci{.029}{.069} \\
    Frequency, 09:00--11:00        & $.45$ \ci{.29}{.57} & $.41$ \ci{.29}{.52} & $.42$ \ci{.32}{.56} & $.44$ \ci{.28}{.57} & $\mathbf{.059}$ \ci{.034}{.101} & $.081$ \ci{.054}{.107} \\
    Frequency, 11:00--13:00        & $.30$ \ci{.23}{.35} & $.32$ \ci{.23}{.39} & $.24$ \ci{.19}{.29} & $.26$ \ci{.22}{.30} & $.051$ \ci{.025}{.096}          & $\mathbf{.034}$ \ci{.018}{.048} \\
    Frequency, 13:00--15:00        & $.20$ \ci{.14}{.25} & $.21$ \ci{.14}{.28} & $.19$ \ci{.13}{.25} & $.22$ \ci{.17}{.27} & $\mathbf{.045}$ \ci{.034}{.067} & $.065$ \ci{.046}{.101} \\
    Frequency, 15:00--17:00        & $.23$ \ci{.19}{.26} & $.23$ \ci{.18}{.29} & $.19$ \ci{.17}{.23} & $.20$ \ci{.15}{.24} & $\mathbf{.019}$ \ci{.012}{.027} & $.032$ \ci{.022}{.042} \\
    \midrule
    Next-activity transition & $.36$ \ci{.29}{.46} & $.35$ \ci{.26}{.45} & $.31$ \ci{.26}{.40} & $.33$ \ci{.27}{.42} & $.107$ \ci{.072}{.146}          & $\mathbf{.091}$ \ci{.085}{.098} \\
    Motifs (3-gram)          & $.61$ \ci{.55}{.67} & $.58$ \ci{.50}{.65} & $.56$ \ci{.49}{.60} & $.59$ \ci{.54}{.66} & $\mathbf{.22}$ \ci{.19}{.27}    & $.22$ \ci{.20}{.25} \\
    Motifs (4-gram)          & $.75$ \ci{.73}{.79} & $.73$ \ci{.68}{.77} & $.72$ \ci{.71}{.74} & $.74$ \ci{.69}{.79} & $\mathbf{.42}$ \ci{.38}{.52}    & $.44$ \ci{.40}{.51} \\
    \bottomrule
  \end{tabular}
  \caption{JS divergence between simulated and real activity
  distributions for all 13 distributional metrics.}
  \label{tab:app-distributional}
\end{table*}

% Originally generated by analysis/09_render_appendix_tables.py, then edited by
% hand (time-of-day labels; activity-count and episode-duration rows dropped
% because Section 7 does not discuss them). Regenerating will revert those
% edits.
% Source: analysis/confidence_intervals_absolute_10000/absolute_metrics_by_method.csv
% Bootstrap: confidence_intervals_absolute_10000 (per-approach), confidence_intervals_v2_10000 (contrasts).
% Method mapping: none = B0, baseline = C1, extracted = C2,
%                 few_shot = C3, markov = C4, hybrid = C5.
%
% Scalar metrics. L1 is the unsigned gap to the participant's real days;
% bias is the signed gap, positive when the simulator overshoots.

\providecommand{\cis}[3]{\begin{tabular}[t]{@{}c@{}}$#1$\\[-3pt]{\scriptsize[#2,\,#3]}\end{tabular}}

\begin{table*}[t]
  \centering
  \scriptsize
  \setlength{\tabcolsep}{3pt}
  \begin{tabular}{lcccccc}
    \toprule
    Metric & B0 & C1 & C2 & C3 & C4 & C5 \\
    \midrule
    \multicolumn{7}{@{}l}{\emph{Unsigned distance ($L_1$)}} \\
    Switches            & \cis{\mathbf{13.0}}{7.6}{17.4}     & \cis{15.2}{10.5}{19.6}            & \cis{15.0}{8.7}{19.3}        & \cis{13.4}{9.4}{17.4}        & \cis{38.0}{24.3}{47.8}    & \cis{32.3}{19.5}{45.3} \\
    Switches, 09:00--11:00    & \cis{3.2}{1.7}{4.9}                & \cis{3.3}{2.3}{4.1}               & \cis{\mathbf{2.9}}{1.9}{4.5} & \cis{3.2}{2.1}{4.6}          & \cis{12.1}{7.9}{18.0}     & \cis{12.8}{8.3}{16.5} \\
    Switches, 11:00--13:00    & \cis{\mathbf{3.1}}{2.0}{4.1}       & \cis{3.8}{2.2}{5.1}               & \cis{3.8}{2.3}{5.2}          & \cis{3.1}{1.3}{4.8}          & \cis{6.2}{2.4}{7.7}       & \cis{4.3}{1.8}{5.8} \\
    Switches, 13:00--15:00    & \cis{\mathbf{3.1}}{2.2}{4.9}       & \cis{3.9}{2.7}{4.9}               & \cis{4.1}{3.0}{6.2}          & \cis{3.4}{2.5}{5.1}          & \cis{8.3}{5.2}{10.9}      & \cis{7.5}{4.7}{12.1} \\
    Switches, 15:00--17:00    & \cis{4.8}{3.3}{7.8}                & \cis{5.4}{4.3}{8.3}               & \cis{5.2}{3.7}{7.4}          & \cis{\mathbf{4.4}}{3.1}{6.1} & \cis{11.8}{8.8}{13.9}     & \cis{6.8}{3.3}{10.1} \\
    \midrule
    \multicolumn{7}{@{}l}{\emph{Signed bias ($\Delta$)}} \\
    Switches            & \cis{-13.0}{-17.4}{-7.6}           & \cis{\mathbf{-10.8}}{-19.1}{-4.3} & \cis{-15.0}{-19.3}{-8.7}     & \cis{-13.4}{-17.4}{-9.4}     & \cis{+38.0}{+24.3}{+47.8} & \cis{+32.3}{+19.5}{+45.3} \\
    Switches, 09:00--11:00    & \cis{-3.1}{-4.8}{-1.1}             & \cis{\mathbf{-.9}}{-3.6}{+1.7}    & \cis{-2.5}{-4.4}{-.4}        & \cis{-2.8}{-4.2}{-.6}        & \cis{+12.1}{+7.9}{+18.0}  & \cis{+12.8}{+8.3}{+16.5} \\
    Switches, 11:00--13:00    & \cis{\mathbf{-2.3}}{-4.1}{+.0}     & \cis{-2.7}{-4.7}{-.8}             & \cis{-3.2}{-5.0}{+.1}        & \cis{-3.1}{-4.8}{-1.3}       & \cis{+5.6}{+1.6}{+7.5}    & \cis{+4.3}{+1.7}{+5.7} \\
    Switches, 13:00--15:00    & \cis{-2.9}{-4.9}{-2.0}             & \cis{\mathbf{-2.9}}{-4.6}{-1.3}   & \cis{-4.1}{-6.2}{-3.0}       & \cis{-3.3}{-5.1}{-2.4}       & \cis{+8.0}{+4.7}{+10.8}   & \cis{+7.4}{+4.7}{+12.4} \\
    Switches, 15:00--17:00    & \cis{-4.7}{-7.8}{-3.3}             & \cis{\mathbf{-4.4}}{-7.8}{-2.8}   & \cis{-5.2}{-7.4}{-3.7}       & \cis{-4.4}{-6.1}{-3.1}       & \cis{+11.8}{+8.8}{+13.9}  & \cis{+6.8}{+3.3}{+10.1} \\
    \bottomrule
  \end{tabular}
  \caption{Activity switches per day, over the full day (09:00--17:00) and within each
  two-hour window. $L_{1}$ is the unsigned gap to the individual's real days
  and $\Delta$ the signed gap, positive when the simulator switches more frequently
  than the real reference.}
  \label{tab:app-scalar}
\end{table*}

\paragraph{Statistical Priors against Persona Conditioning.}
% \label{app:results-priors}
Tables~\ref{tab:app-prior-day-local}--\ref{tab:app-prior-scalar} report the eight contrasts of each statistical-prior approach (C4, C5) against each persona-based approach and the control (B0--C3), which are the contrasts \S\ref{result:stat-activity-sequence} and \S\ref{result:fragmented} count.

% Originally generated by analysis/09_render_appendix_tables.py, then edited by
% hand (asterisks dropped since every interval excludes zero). Regenerating
% will revert those edits.
% Source: analysis/confidence_intervals_v2_10000/divergence_3_gram_persona.csv
% Source: analysis/confidence_intervals_v2_10000/divergence_4_gram_persona.csv
% Source: analysis/confidence_intervals_v2_10000/divergence_activity_frequency_distribution_persona.csv
% Source: analysis/confidence_intervals_v2_10000/divergence_activity_time_distribution_persona.csv
% Source: analysis/confidence_intervals_v2_10000/divergence_transition_persona.csv
% Bootstrap: confidence_intervals_absolute_10000 (per-approach), confidence_intervals_v2_10000 (contrasts).
% Method mapping: none = B0, baseline = C1, extracted = C2,
%                 few_shot = C3, markov = C4, hybrid = C5.
%
% Backs the claim in Section 7.1 that all eight contrasts exclude zero on
% each of these five metrics.
%
% Contrasts are oriented prior minus persona, so a negative estimate means
% the statistical prior is closer to real behaviour.

\providecommand{\ci}[2]{{\scriptsize\,[#1,\,#2]}}

\begin{table*}[t]
  \centering
  \scriptsize
  \setlength{\tabcolsep}{2.5pt}
  \begin{tabular}{lccccc}
    \toprule
    Contrast & Time-allocation & Frequency & Transition & Motifs (3-gram) & Motifs (4-gram) \\
    \midrule
    C4 vs.\ B0 & $-.204$ \ci{-.272}{-.122} & $-.158$ \ci{-.192}{-.101} & $-.256$ \ci{-.442}{-.164} & $-.39$ \ci{-.46}{-.21} & $-.33$ \ci{-.40}{-.17} \\
    C4 vs.\ C1 & $-.235$ \ci{-.328}{-.132} & $-.168$ \ci{-.213}{-.115} & $-.239$ \ci{-.354}{-.066} & $-.36$ \ci{-.46}{-.21} & $-.30$ \ci{-.39}{-.20} \\
    C4 vs.\ C2 & $-.179$ \ci{-.269}{-.097} & $-.124$ \ci{-.157}{-.085} & $-.208$ \ci{-.286}{-.087} & $-.34$ \ci{-.41}{-.14} & $-.30$ \ci{-.36}{-.15} \\
    C4 vs.\ C3 & $-.212$ \ci{-.297}{-.126} & $-.150$ \ci{-.191}{-.100} & $-.222$ \ci{-.397}{-.115} & $-.37$ \ci{-.45}{-.20} & $-.32$ \ci{-.41}{-.22} \\
    \midrule
    C5 vs.\ B0 & $-.203$ \ci{-.274}{-.119} & $-.148$ \ci{-.182}{-.093} & $-.272$ \ci{-.401}{-.170} & $-.39$ \ci{-.45}{-.24} & $-.32$ \ci{-.39}{-.20} \\
    C5 vs.\ C1 & $-.235$ \ci{-.324}{-.128} & $-.158$ \ci{-.199}{-.103} & $-.255$ \ci{-.387}{-.130} & $-.36$ \ci{-.45}{-.23} & $-.29$ \ci{-.39}{-.22} \\
    C5 vs.\ C2 & $-.179$ \ci{-.271}{-.094} & $-.115$ \ci{-.145}{-.082} & $-.224$ \ci{-.370}{-.145} & $-.34$ \ci{-.39}{-.22} & $-.29$ \ci{-.34}{-.16} \\
    C5 vs.\ C3 & $-.212$ \ci{-.300}{-.123} & $-.141$ \ci{-.176}{-.099} & $-.237$ \ci{-.382}{-.154} & $-.37$ \ci{-.45}{-.26} & $-.30$ \ci{-.42}{-.24} \\
    \bottomrule
  \end{tabular}
  \caption{Paired contrasts of the statistical priors (C4--C5) against the
  persona-based approaches and the control (B0--C3) for day-level (time
  allocation, frequency) and local metrics (transitions, 3-gram, 4-gram).
  Estimates are differences in JS divergence oriented as named. The negative values mean the statistical prior is closer to real behavior. Every interval in the table excludes zero.}
  \label{tab:app-prior-day-local}
\end{table*}

% Originally generated by analysis/09_render_appendix_tables.py, then edited by
% hand (windows laid out side by side; asterisks dropped since every interval
% excludes zero). Regenerating will revert those edits.
% Source: analysis/confidence_intervals_v2_10000/divergence_time_activity_time_distribution_persona.csv
% Source: analysis/confidence_intervals_v2_10000/divergence_time_activity_frequency_distribution_persona.csv
% Bootstrap: confidence_intervals_absolute_10000 (per-approach), confidence_intervals_v2_10000 (contrasts).
% Method mapping: none = B0, baseline = C1, extracted = C2,
%                 few_shot = C3, markov = C4, hybrid = C5.
%
% Backs the claim in Section 7.1 that the time-allocation contrasts weaken
% over the course of the day while the activity-frequency contrasts hold in
% every window.

\providecommand{\ci}[2]{{\scriptsize\,[#1,\,#2]}}

\begin{table*}[t]
  \centering
  \footnotesize
  \setlength{\tabcolsep}{4pt}
  \begin{tabular}{lcc@{\hspace{2.5em}}lcc}
    \toprule
    Contrast & Time-allocation & Frequency & Contrast & Time-allocation & Frequency \\
    \midrule
    \multicolumn{3}{@{}l}{\emph{09:00--11:00}} & \multicolumn{3}{@{}l}{\emph{13:00--15:00}} \\
    C4 vs.\ B0 & $-.418$ \ci{-.732}{-.166} & $-.395$ \ci{-.514}{-.217} & C4 vs.\ B0 & $-.124$ \ci{-.188}{-.048} & $-.155$ \ci{-.222}{-.074} \\
    C4 vs.\ C1 & $-.397$ \ci{-.693}{-.145} & $-.355$ \ci{-.488}{-.200} & C4 vs.\ C1 & $-.110$ \ci{-.161}{-.038} & $-.168$ \ci{-.241}{-.070} \\
    C4 vs.\ C2 & $-.383$ \ci{-.706}{-.162} & $-.363$ \ci{-.504}{-.262} & C4 vs.\ C2 & $-.091$ \ci{-.197}{-.035} & $-.141$ \ci{-.221}{-.075} \\
    C4 vs.\ C3 & $-.413$ \ci{-.737}{-.148} & $-.382$ \ci{-.521}{-.185} & C4 vs.\ C3 & $-.118$ \ci{-.179}{-.071} & $-.176$ \ci{-.219}{-.092} \\
    C5 vs.\ B0 & $-.402$ \ci{-.674}{-.165} & $-.373$ \ci{-.520}{-.193} & C5 vs.\ B0 & $-.132$ \ci{-.179}{-.089} & $-.135$ \ci{-.189}{-.076} \\
    C5 vs.\ C1 & $-.382$ \ci{-.635}{-.144} & $-.333$ \ci{-.493}{-.183} & C5 vs.\ C1 & $-.118$ \ci{-.157}{-.061} & $-.148$ \ci{-.214}{-.060} \\
    C5 vs.\ C2 & $-.368$ \ci{-.648}{-.157} & $-.341$ \ci{-.511}{-.229} & C5 vs.\ C2 & $-.099$ \ci{-.186}{-.061} & $-.121$ \ci{-.218}{-.079} \\
    C5 vs.\ C3 & $-.398$ \ci{-.679}{-.139} & $-.360$ \ci{-.508}{-.162} & C5 vs.\ C3 & $-.126$ \ci{-.180}{-.103} & $-.156$ \ci{-.200}{-.116} \\
    \midrule
    \multicolumn{3}{@{}l}{\emph{11:00--13:00}} & \multicolumn{3}{@{}l}{\emph{15:00--17:00}} \\
    C4 vs.\ B0 & $-.336$ \ci{-.475}{-.239} & $-.250$ \ci{-.315}{-.178} & C4 vs.\ B0 & $-.124$ \ci{-.201}{-.046} & $-.206$ \ci{-.257}{-.151} \\
    C4 vs.\ C1 & $-.369$ \ci{-.543}{-.220} & $-.268$ \ci{-.334}{-.177} & C4 vs.\ C1 & $-.167$ \ci{-.294}{-.105} & $-.206$ \ci{-.303}{-.132} \\
    C4 vs.\ C2 & $-.269$ \ci{-.416}{-.168} & $-.193$ \ci{-.251}{-.156} & C4 vs.\ C2 & $-.101$ \ci{-.163}{-.043} & $-.171$ \ci{-.239}{-.141} \\
    C4 vs.\ C3 & $-.314$ \ci{-.463}{-.171} & $-.213$ \ci{-.260}{-.171} & C4 vs.\ C3 & $-.130$ \ci{-.190}{-.031} & $-.182$ \ci{-.236}{-.114} \\
    C5 vs.\ B0 & $-.336$ \ci{-.488}{-.241} & $-.267$ \ci{-.318}{-.192} & C5 vs.\ B0 & $-.111$ \ci{-.182}{-.034} & $-.193$ \ci{-.234}{-.150} \\
    C5 vs.\ C1 & $-.368$ \ci{-.601}{-.229} & $-.285$ \ci{-.360}{-.163} & C5 vs.\ C1 & $-.154$ \ci{-.270}{-.094} & $-.193$ \ci{-.292}{-.133} \\
    C5 vs.\ C2 & $-.268$ \ci{-.423}{-.170} & $-.210$ \ci{-.248}{-.156} & C5 vs.\ C2 & $-.088$ \ci{-.141}{-.041} & $-.158$ \ci{-.215}{-.141} \\
    C5 vs.\ C3 & $-.314$ \ci{-.492}{-.181} & $-.230$ \ci{-.259}{-.168} & C5 vs.\ C3 & $-.117$ \ci{-.171}{-.045} & $-.170$ \ci{-.214}{-.109} \\
    \bottomrule
  \end{tabular}
  \caption{Paired contrasts of the statistical priors (C4--C5) against the
  persona-based approaches and the control (B0--C3), within each two-hour
  window. Estimates are differences in JS divergence oriented as named.
  The negative values mean the statistical prior is closer to real
  behavior. Every interval in the table excludes zero.}
  \label{tab:app-prior-tod}
\end{table*}

% Originally generated by analysis/09_render_appendix_tables.py, then edited by
% hand (activity-count and episode-duration columns dropped because Section 7
% does not discuss them; asterisks and redundant + signs removed, both noted in
% the caption instead). Regenerating will revert those edits.
% Source: analysis/confidence_intervals_v2_10000/divergence_activity_switch_count_pairwise.csv
% Bootstrap: confidence_intervals_absolute_10000 (per-approach), confidence_intervals_v2_10000 (contrasts).
% Method mapping: none = B0, baseline = C1, extracted = C2,
%                 few_shot = C3, markov = C4, hybrid = C5.
%
% Backs the claim in Section 7.2 on activity switches.
%
% Contrasts are oriented prior minus persona.

\providecommand{\ci}[2]{{\scriptsize\,[#1,\,#2]}}

\begin{table*}[t]
  \centering
  \small
  \setlength{\tabcolsep}{4pt}
  \begin{tabular}{lc}
    \toprule
    Contrast & Switches \\
    \midrule
    C4 vs.\ B0 & $24.9$ \ci{13.5}{42.6} \\
    C4 vs.\ C1 & $22.8$ \ci{9.1}{37.5}  \\
    C4 vs.\ C2 & $22.9$ \ci{11.6}{37.8} \\
    C4 vs.\ C3 & $24.6$ \ci{10.8}{41.6} \\
    \midrule
    C5 vs.\ B0 & $19.3$ \ci{8.8}{37.4}  \\
    C5 vs.\ C1 & $17.1$ \ci{5.2}{34.3}  \\
    C5 vs.\ C2 & $17.3$ \ci{7.2}{37.6}  \\
    C5 vs.\ C3 & $18.9$ \ci{6.8}{38.4}  \\
    \bottomrule
  \end{tabular}
  \caption{Paired contrasts of the statistical priors (C4--C5) against the
  persona-based approaches and the control (B0--C3) on the activity-switch
  gap. Estimates are differences in $L_{1}$ switches per day, oriented as named. Every estimate and every interval bound is positive, which indicates the statistical priors are further from the real switch rate.
  Every interval in the table
  excludes zero.}
  \label{tab:app-prior-scalar}
\end{table*}

\paragraph{Within-person Variability.}
% \label{app:results-variability}
\autoref{tab:app-variability} reports the within-person session-to-session
variability behind \S\ref{result:fragmented}, both as each approach's own
level and as all 15 paired contrasts. 

% Originally generated by analysis/09_render_appendix_tables.py, then edited by
% hand (per-approach and contrast column labels split into two header rows).
% Regenerating will revert those edits.
% Source: analysis/confidence_intervals_absolute_10000/absolute_metrics_by_method.csv
% Source: analysis/confidence_intervals_v2_10000/divergence_within_pid_pairwise.csv
% Bootstrap: confidence_intervals_absolute_10000 (per-approach), confidence_intervals_v2_10000 (contrasts).
% Method mapping: none = B0, baseline = C1, extracted = C2,
%                 few_shot = C3, markov = C4, hybrid = C5.
%
% Backs the claim in Section 7.2 that the statistical priors produce the
% most homogeneous days. $V$ is the mean pairwise JS between two sessions of
% the same individual; $L_1(\mathrm{JS})$ is the unsigned gap between an
% approach's $V$ and the real one.
%
% All 15 approach pairs are listed, not only the eight prior-vs-persona
% contrasts, because Section 7.2 reads the persona-persona pairs too.

\providecommand{\ci}[2]{{\scriptsize\,[#1,\,#2]}}

\begin{table*}[t]
  \centering
  \footnotesize
  \setlength{\tabcolsep}{6pt}
  \begin{tabular}{lcc}
    \toprule
    & $V^{\mathrm{sim}}$ & $L_{1}(\mathrm{JS})$ \\
    \midrule
    \multicolumn{3}{@{}l}{\emph{Per approach}} \\
    B0 No-Persona    & $.058$ \ci{.039}{.066}                     & $.043$ \ci{.027}{.060} \\
    C1 Persona-Auth. & $.043$ \ci{.036}{.053}                     & $.054$ \ci{.038}{.066} \\
    C2 Persona-Inf.  & $.064$ \ci{.061}{.069}                     & $.045$ \ci{.026}{.064} \\
    C3 \,+ Few-Shot  & $.044$ \ci{.036}{.050}                     & $.054$ \ci{.037}{.088} \\
    C4 Stat.-Prior   & $.018$ \ci{.014}{.027}                     & $.079$ \ci{.051}{.104} \\
    C5 Hybrid        & $.026$ \ci{.019}{.043}                     & $.070$ \ci{.037}{.093} \\
    Real             & $.097$ \ci{.077}{.119}                     & --- \\
    \midrule
    & $\Delta V$ & $\Delta L_{1}(\mathrm{JS})$ \\
    \midrule
    \multicolumn{3}{@{}l}{\emph{Pairwise contrasts (row $-$ column arm)}} \\
    
    B0 vs.\ C1      & $+.015$ \ci{-.013}{+.028}                  & $-.010$ \ci{-.032}{+.008} \\
    B0 vs.\ C2       & $-.007$ \ci{-.027}{+.003}                  & $-.002$ \ci{-.047}{+.022} \\
    B0 vs.\ C3       & $+.013$ \ci{-.008}{+.036}                  & $-.011$ \ci{-.036}{+.010} \\
    B0 vs.\ C4       & \textbf{$\ast$}\,$+.040$ \ci{+.021}{+.054} & \textbf{$\ast$}\,$-.036$ \ci{-.054}{-.007} \\
    B0 vs.\ C5       & \textbf{$\ast$}\,$+.031$ \ci{+.009}{+.048} & $-.027$ \ci{-.048}{+.015} \\
    C1 vs.\ C2       & \textbf{$\ast$}\,$-.021$ \ci{-.036}{-.005} & $+.009$ \ci{-.013}{+.021} \\
    C1 vs.\ C3       & $-.001$ \ci{-.013}{+.028}                  & $-.000$ \ci{-.028}{+.009} \\
    C1 vs.\ C4       & \textbf{$\ast$}\,$+.026$ \ci{+.008}{+.041} & \textbf{$\ast$}\,$-.025$ \ci{-.042}{-.007} \\
    C1 vs.\ C5       & $+.017$ \ci{-.014}{+.033}                  & $-.016$ \ci{-.033}{+.014} \\
    C2 vs.\ C3       & \textbf{$\ast$}\,$+.020$ \ci{+.011}{+.031} & $-.009$ \ci{-.024}{+.023} \\
    C2 vs.\ C4       & \textbf{$\ast$}\,$+.047$ \ci{+.030}{+.055} & \textbf{$\ast$}\,$-.034$ \ci{-.048}{-.006} \\
    C2 vs.\ C5       & \textbf{$\ast$}\,$+.038$ \ci{+.012}{+.049} & \textbf{$\ast$}\,$-.025$ \ci{-.042}{-.001} \\
    C3 vs.\ C4       & \textbf{$\ast$}\,$+.027$ \ci{+.017}{+.035} & \textbf{$\ast$}\,$-.025$ \ci{-.035}{-.011} \\
    C3 vs.\ C5       & \textbf{$\ast$}\,$+.018$ \ci{+.002}{+.027} & $-.016$ \ci{-.027}{+.009} \\
    C4 vs.\ C5       & \textbf{$\ast$}\,$-.009$ \ci{-.018}{-.006} & \textbf{$\ast$}\,$+.009$ \ci{+.006}{+.018} \\
    \bottomrule
  \end{tabular}
  \caption{Within-person session-to-session variability.
  The upper block gives each approach's own level; the lower block gives all 15 paired contrasts, oriented as named, so a negative $\Delta V$
  means the first arm produced the more uniform days. $\ast$ marks intervals
  excluding zero.}
  \label{tab:app-variability}
\end{table*}

\paragraph{Contrasts among Persona-based Approaches.}
% \label{app:results-persona}
\autoref{tab:app-persona} reports the six pairwise contrasts among B0--C3 for each of the 13 distributional metrics, which is referenced in \S\ref{result:persona-conditioning}.

% Originally generated by analysis/09_render_appendix_tables.py, then edited by
% hand (time-of-day labels). Regenerating will revert those edits.
% Source: analysis/confidence_intervals_v2_10000/divergence_3_gram_persona.csv
% Source: analysis/confidence_intervals_v2_10000/divergence_4_gram_persona.csv
% Source: analysis/confidence_intervals_v2_10000/divergence_activity_frequency_distribution_persona.csv
% Source: analysis/confidence_intervals_v2_10000/divergence_activity_time_distribution_persona.csv
% Source: analysis/confidence_intervals_v2_10000/divergence_time_activity_frequency_distribution_persona.csv
% Source: analysis/confidence_intervals_v2_10000/divergence_time_activity_time_distribution_persona.csv
% Source: analysis/confidence_intervals_v2_10000/divergence_transition_persona.csv
% Bootstrap: confidence_intervals_absolute_10000 (per-approach), confidence_intervals_v2_10000 (contrasts).
% Method mapping: none = B0, baseline = C1, extracted = C2,
%                 few_shot = C3, markov = C4, hybrid = C5.
%
% Backs the claim in Section 7.3. Six contrasts per metric across 13
% distributional metrics gives 78 contrasts; the ones marked with an
% asterisk are those whose interval excludes zero.

\providecommand{\cis}[3]{\begin{tabular}[t]{@{}c@{}}$#1$\\[-3pt]{\scriptsize[#2,\,#3]}\end{tabular}}

\begin{table*}[t]
  \centering
  \scriptsize
  \setlength{\tabcolsep}{1.5pt}
  \begin{tabular}{lcccccc}
    \toprule
    Metric & B0--C1 & B0--C2 & B0--C3 & C1--C2 & C1--C3 & C2--C3 \\
    \midrule
    Time-allocation          & $\ast$\cis{-.031}{-.060}{-.010} & $\ast$\cis{+.025}{+.006}{+.047} & \cis{-.008}{-.036}{+.010} & $\ast$\cis{+.056}{+.031}{+.075} & $\ast$\cis{+.023}{+.012}{+.032} & $\ast$\cis{-.033}{-.045}{-.017} \\
    Frequency                & \cis{-.010}{-.032}{+.043}       & $\ast$\cis{+.034}{+.011}{+.049} & \cis{+.008}{-.007}{+.038} & \cis{+.044}{-.024}{+.067}       & \cis{+.017}{-.038}{+.051}       & $\ast$\cis{-.026}{-.061}{-.011} \\
    \midrule
    Time-allocation, 09:00--11:00  & $\ast$\cis{+.021}{+.000}{+.046} & $\ast$\cis{+.034}{+.001}{+.100} & \cis{+.005}{-.029}{+.043} & \cis{+.014}{-.042}{+.086}       & \cis{-.016}{-.065}{+.039}       & \cis{-.029}{-.073}{+.024} \\
    Time-allocation, 11:00--13:00  & \cis{-.032}{-.084}{+.048}       & $\ast$\cis{+.067}{+.058}{+.089} & \cis{+.022}{-.029}{+.086} & $\ast$\cis{+.100}{+.035}{+.157} & $\ast$\cis{+.055}{+.024}{+.080} & \cis{-.045}{-.099}{+.001} \\
    Time-allocation, 13:00--15:00  & \cis{+.014}{-.008}{+.043}       & \cis{+.033}{-.000}{+.059}       & \cis{+.006}{-.016}{+.032} & \cis{+.019}{-.029}{+.063}       & \cis{-.007}{-.053}{+.032}       & \cis{-.027}{-.060}{+.001} \\
    Time-allocation, 15:00--17:00  & $\ast$\cis{-.043}{-.085}{-.006} & \cis{+.023}{-.020}{+.046}       & \cis{-.006}{-.102}{+.024} & $\ast$\cis{+.066}{+.049}{+.112} & \cis{+.037}{-.023}{+.088}       & \cis{-.029}{-.072}{+.001} \\
    Frequency, 09:00--11:00        & $\ast$\cis{+.040}{+.012}{+.081} & \cis{+.032}{-.033}{+.097}       & \cis{+.013}{-.061}{+.082} & \cis{-.008}{-.096}{+.072}       & \cis{-.027}{-.118}{+.063}       & \cis{-.019}{-.164}{+.046} \\
    Frequency, 11:00--13:00        & \cis{-.018}{-.081}{+.077}       & $\ast$\cis{+.057}{+.025}{+.082} & \cis{+.037}{-.047}{+.080} & $\ast$\cis{+.075}{+.006}{+.120} & \cis{+.055}{-.003}{+.109}       & \cis{-.020}{-.070}{+.000} \\
    Frequency, 13:00--15:00        & \cis{-.013}{-.030}{+.023}       & \cis{+.014}{-.022}{+.107}       & \cis{-.021}{-.049}{+.036} & \cis{+.027}{-.014}{+.124}       & \cis{-.007}{-.075}{+.040}       & \cis{-.035}{-.063}{+.004} \\
    Frequency, 15:00--17:00        & \cis{+.000}{-.058}{+.034}       & $\ast$\cis{+.035}{+.011}{+.065} & \cis{+.024}{-.020}{+.110} & \cis{+.035}{-.010}{+.082}       & \cis{+.023}{-.056}{+.112}       & \cis{-.012}{-.038}{+.046} \\
    \midrule
    Next-activity transition & \cis{+.017}{-.060}{+.109}       & \cis{+.048}{-.016}{+.195}       & \cis{+.035}{-.003}{+.072} & \cis{+.031}{-.014}{+.094}       & \cis{+.017}{-.049}{+.112}       & \cis{-.014}{-.162}{+.052} \\
    Motifs (3-gram)          & $\ast$\cis{+.035}{+.011}{+.109} & \cis{+.049}{-.012}{+.069}       & \cis{+.020}{-.007}{+.051} & \cis{+.014}{-.122}{+.056}       & \cis{-.014}{-.073}{+.019}       & \cis{-.028}{-.068}{+.049} \\
    Motifs (4-gram)          & \cis{+.029}{-.006}{+.061}       & $\ast$\cis{+.030}{+.003}{+.053} & \cis{+.011}{-.022}{+.056} & \cis{+.001}{-.069}{+.040}       & $\ast$\cis{-.017}{-.038}{-.002} & \cis{-.019}{-.063}{+.052} \\
    \bottomrule
  \end{tabular}
  \caption{Paired contrasts among the no-persona control and the
  persona-based approaches (B0--C3), for all 13 distributional metrics. Estimates are differences in JS divergence oriented as
  named. $\ast$ marks intervals excluding zero.}
  \label{tab:app-persona}
\end{table*}

\paragraph{Individual versus Population Aggregation.}
% \label{app:results-aggregation}

\autoref{tab:app-ind-pop} reports each of the 13 distributional metrics
computed at the individual level beside the same metric computed after
averaging distributions over the population, together with the pooled
difference behind \S\ref{result:level-of-analysis}. 

% Originally generated by analysis/09_render_appendix_tables.py, then edited by
% hand (time-of-day labels; asterisks dropped since every interval excludes
% zero). Regenerating will revert those edits.
% Source: analysis/confidence_intervals_absolute_10000/absolute_metrics_by_method.csv
% Source: analysis/confidence_intervals_v2_10000/individual_vs_population_pooled.csv
% Bootstrap: confidence_intervals_absolute_10000 (per-approach), confidence_intervals_v2_10000 (contrasts).
% Method mapping: none = B0, baseline = C1, extracted = C2,
%                 few_shot = C3, markov = C4, hybrid = C5.
%
% Backs the claim in Section 7.4. The individual and population columns are
% means over the six approaches of the per-approach estimates; the contrast
% is the pooled individual-minus-population difference, which averages the
% gap over approaches within each bootstrap replicate.

\providecommand{\ci}[2]{{\scriptsize\,[#1,\,#2]}}

\begin{table*}[t]
  \centering
  \footnotesize
  \setlength{\tabcolsep}{6pt}
  \begin{tabular}{lccc}
    \toprule
    Metric & Individual & Population & $\Delta$ (ind.\ $-$ pop.) \\
    \midrule
    Time-allocation          & $.148$ & $.128$ & $+.020$ \ci{+.011}{+.023} \\
    Frequency                & $.119$ & $.100$ & $+.019$ \ci{+.014}{+.019} \\
    \midrule
    Time-allocation, 09:00--11:00  & $.293$ & $.242$ & $+.052$ \ci{+.020}{+.069} \\
    Time-allocation, 11:00--13:00  & $.245$ & $.198$ & $+.048$ \ci{+.034}{+.053} \\
    Time-allocation, 13:00--15:00  & $.123$ & $.087$ & $+.036$ \ci{+.034}{+.038} \\
    Time-allocation, 15:00--17:00  & $.117$ & $.091$ & $+.026$ \ci{+.019}{+.029} \\
    Frequency, 09:00--11:00        & $.312$ & $.262$ & $+.050$ \ci{+.047}{+.050} \\
    Frequency, 11:00--13:00        & $.202$ & $.151$ & $+.052$ \ci{+.048}{+.052} \\
    Frequency, 13:00--15:00        & $.155$ & $.112$ & $+.043$ \ci{+.033}{+.045} \\
    Frequency, 15:00--17:00        & $.149$ & $.122$ & $+.027$ \ci{+.021}{+.028} \\
    \midrule
    Next-activity transition & $.258$ & $.189$ & $+.069$ \ci{+.040}{+.085} \\
    Motifs (3-gram)          & $.465$ & $.374$ & $+.091$ \ci{+.084}{+.091} \\
    Motifs (4-gram)          & $.634$ & $.520$ & $+.115$ \ci{+.108}{+.115} \\
    \bottomrule
  \end{tabular}
  \caption{JS divergence computed at the individual level against the same
  quantity computed after averaging distributions over the population, for all
  13 distributional metrics. A positive $\Delta$ means population-level
  aggregation reports the smaller divergence. Every interval in the table
  excludes zero.}
  \label{tab:app-ind-pop}
\end{table*}

% The per-backbone breakdown belongs in Appendix F
% (sections/appendix/backbone.tex), which already has its own prose and whose
% Table 6 is reproduced by the day-level block of:
% \input{sections/appendix/additional-data-results/F_backbones}

\section{AI Usage Disclosure}
We used Claude (Anthropic) to polish the writing of this manuscript and to
assist in implementing our experimental and analysis infrastructure. 
The authors have reviewed and verified all
AI-assisted outputs and take full responsibility for the accuracy and
originality of the contents of this work.

\end{document}